\documentclass[pdflatex,sn-mathphys-num]{sn-jnl}

\usepackage{graphicx}%
\usepackage{multirow}%
\usepackage{amsmath,amssymb,amsfonts}%
\usepackage{amsthm}%
\usepackage{mathrsfs}%
\usepackage[title]{appendix}%
\usepackage{xcolor}%

\usepackage{textcomp}%
\usepackage{manyfoot}%
\usepackage{booktabs}%
\usepackage{algorithm}%
\usepackage{algorithmicx}%
\usepackage{algpseudocode}%
\usepackage{listings}%
\usepackage{natbib}%
\usepackage{multirow}%
\usepackage{subcaption}%

\usepackage{tabularx}
\usepackage{caption}
\usepackage{siunitx}
\usepackage[section]{placeins}  
\DeclareMathOperator*{\argmax}{arg\,max}

\theoremstyle{thmstyleone}%
\theoremstyle{thmstyletwo}%

\theoremstyle{thmstylethree}%

\begin{document}

\title[Article Title]{
Designing faster mixed integer linear programming algorithm via learning the optimal path
}


\author[1,2]{\fnm{Ruizhi} \sur{Liu}}\email{liuruizhi19s@ict.ac.cn}
\equalcont{These authors contributed equally to this work.}

\author[1,2]{\fnm{Liming} \sur{Xu}}\email{limingxu0713@gmail.com}
\equalcont{These authors contributed equally to this work.}

\author[1,2]{\fnm{Xulin} \sur{Huang}}\email{huangxulin@gs.zzu.edu.cn}
\equalcont{These authors contributed equally to this work.}

\author[1,2,3]{\fnm{Jingyan} \sur{Sui}}\email{suijingyan18@mails.ucas.ac.cn}

\author[1,2]{\fnm{Shizhe} \sur{Ding}}\email{dingshizhe15@mails.ucas.ac.cn}

\author[1,2]{\fnm{Boyang} \sur{Xia}}\email{xiaboyang20@mails.ucas.ac.cn}

\author[1,2,4]{\fnm{Chungong} \sur{Yu}}\email{yuchungong@ict.ac.cn}

\author*[1,2,4]{\fnm{Dongbo} \sur{Bu}}\email{dbu@ict.ac.cn}

\affil[1]{\orgdiv{SKLP, Institute of Computing Technology}, \orgname{Chinese Academy of Sciences}, \orgaddress{\city{Beijing}, \postcode{100190}, \country{China}}}
\affil[2]{\orgdiv{University of Chinese Academy of Science}, \orgaddress{\city{Beijing}, \postcode{100049}, \country{China}}}
\affil[3]{\orgdiv{School of Computer Science}, \orgname{Liaocheng University}, \orgaddress{\city{Liaocheng}, \postcode{252000}, \country{China}}}

\affil[4]{\orgdiv{Central China Research Institute for Artificial Intelligence Technologies}, \orgname{Henan Academy of Sciences}, \orgaddress{\city{Zhengzhou}, \postcode{450046}, \country{China}}}

\abstract{
Designing faster algorithms for solving Mixed-Integer Linear Programming (MILP) problems is highly desired across numerous practical domains, as a vast array of complex real-world challenges can be effectively modeled as MILP formulations. Solving these problems typically employs the branch-and-bound algorithm, the core of which can be conceived as searching for a path of nodes (or sub-problems) that contains the optimal solution to the original MILP problem. Traditional approaches to finding this path rely heavily on hand-crafted, intuition-based heuristic strategies, which often suffer from unstable and unpredictable performance across different MILP problem instances. To address this limitation, we introduce DeepBound, a deep learning-based node selection algorithm that automates the learning of such human intuition from data. The core of DeepBound lies in learning to prioritize nodes containing the optimal solution, thereby improving solving efficiency. DeepBound introduces a multi-level feature fusion network to capture the node representations. To tackle the inherent node imbalance in branch-and-bound trees, DeepBound employs a pairwise training paradigm that enhances the model's ability to discriminate between nodes. Extensive experiments on three NP-hard MILP benchmarks demonstrate that DeepBound achieves superior solving efficiency over conventional heuristic rules and existing learning-based approaches, obtaining optimal feasible solutions with significantly reduced computation time. Moreover, DeepBound demonstrates strong generalization capability on large and complex instances. The analysis of its learned features reveals that the method can automatically discover more flexible and robust feature selection, which may effectively improve and potentially replace human-designed heuristic rules.
}




\maketitle
\clearpage

\section{Introduction}\label{sec1}

Mixed-Integer Linear Programming (MILP) serves as a fundamental mathematical modeling tool to address complex real-world combinatorial optimization challenges, including vehicle routing \cite{louati2021mixed, yang2025adaptive, pedram2023incorporating},  integrated circuit design \cite{cheng2025so3, liu2024robust}, logistics operations\cite{baller2022optimizing, rajak2022multi} and resource allocation \cite{ahmadian2023optimal, lippi2021mixed, nematian2023two}. However, the inherent complexity of general MILP problems is classified as NP-hard, resulting in exponential growth in computational time as problem dimensions increase. The branch-and-bound algorithm remains the cornerstone for solving MILP problems \cite{bengio2021machine, gleixner2018scip}, relying on recursively partitioning the solution space into a search tree. This method systematically explores the optimal solution at the leaf nodes of the tree, which has been extensively studied and refined over decades \cite{benichou1971experiments, achterberg2013mixed, dakin1965tree}. Various hand-crafted heuristic strategies have been developed to guide node and variable selection, thereby controlling the size of branching tree and accelerating the solving process. These approaches are typically based on redetermined single- or multi-criteria metrics extracted from branching tree, which introduces limitations that their effectiveness often varies significantly across diverse MILP problems.

\subsubsection*{MILP formulation and branch-and-bound algorithm}
Combinatorial optimization problems, e.g., the set covering problem, combinatorial auction problem and capacitated facility location problem depicted in Figure 1.b, can be written in the general MILP form in Figure 1.a \cite{achterberg2009scip}, where $\mathbf{c} \in \mathbb{R}^n$ is the objective coefficient vector that is commonly used to describe the cost function of the original problem. $\mathbf{A} \in \mathbb{R}^{m \times n}$ and $\mathbf{b} \in \mathbb{R}^m$ represent the constraint coefficient matrix and the constraint right-hand-side vector, respectively, and the size of a MILP problem is typically measured by the number of rows ($m$) and columns ($n$) of the constraint matrix $\mathbf{A}$. 

When solving the MILP problem, such as the $2000\times1000$ set covering problem in Figure 1.c, the branch-and-bound algorithm will first solve the linear programming relaxation of the original MILP problem, then partition the larger MILP problem into two smaller sub-problems on the variable that does not respect integrality \cite{achterberg2009scip}. By executing such binary decomposition iteratively, the algorithm yields a search tree, as shown in Figure 1.d, and searches for the optimal solution to the original MILP problem on newly generated leaf nodes. The nodes containing the optimal solution in their solution space are labeled in red, which form an \textbf{optimal path} from the root node to the node where the optimal solution was found in the branch-and-bound search tree\cite{he2014learning}.

\subsubsection*{Learning-based enhancement of the branch-and-bound algorithm}
Learning-based approaches have shown promising results in accelerating or even replacing human-designed heuristic rules within the branch-and-bound algorithm. The problem of branching variable selection can be framed as a ranking problem, where the model was trained to learn the order of branching variables directly from the outcomes of strong branching rule \cite{khalil2016learning}, or to assign scores to each candidate branching variable \cite{alvarez2017machine}. 

Graph neural networks (GNNs) was used to encode the relationships between variables and constraints in each sub-problem, allowing the model to effectively learn the variable selection of the strong branching rule \cite{gasse2019exact, nair2020solving} or to discover improved variable selection orders through reinforcement learning \cite{qu2022improved, parsonson2023reinforcement}. Furthermore, Ding et al. \cite{ding2020accelerating} introduced a GNN-based learning framework to explore primal heuristics during the MILP solving process, leveraging tripartite graph representations of nodes to learn how to generate high-quality feasible solutions. Similarly, GNN-based models were also introduced to efficiently explore the neighborhood of an initial feasible solution, aiming to identify better solutions \cite{nair2020solving, paulus2023learning, liu2025apollomilpalternatingpredictioncorrectionneural}. Newer work also integrated graph neural architecture search algorithms to automatically find the best GNNs for a given NP-hard combinatorial optimization problem \cite{liu2024combinatorialoptimizationautomatedgraph}. Machine learning has also been successfully applied to discover better cutting plane addition sequences and apply the learned models to accelerate the solving process of different MILP problems \cite{tang2020reinforcement, huang2022learning, paulus2022learning, deza2023machine}. 

\begin{figure}[H]
\centering
\includegraphics[width=0.8\columnwidth]{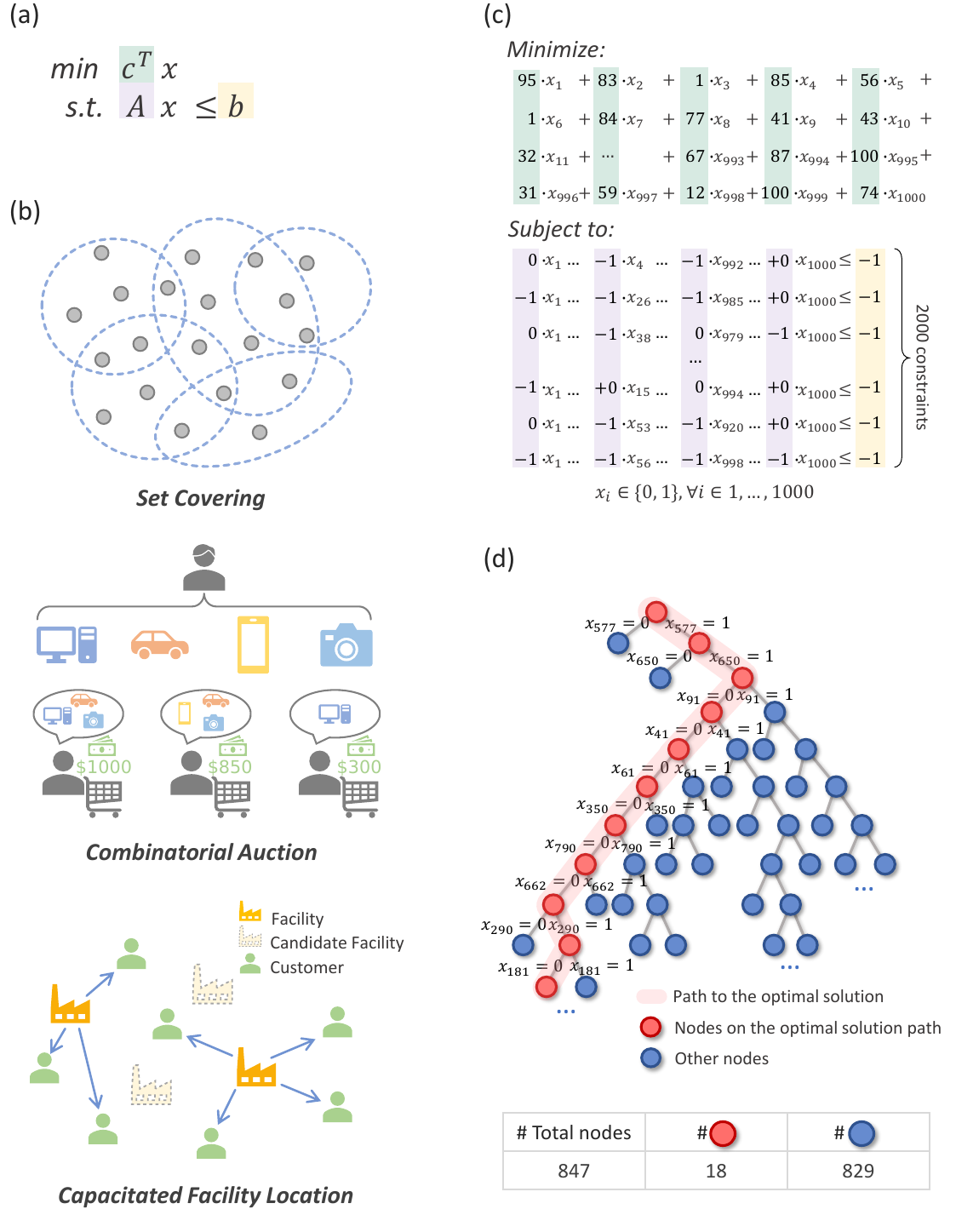}
\caption{\textbf{MILP problems and branch-and-bound tree}. \textbf{a,} The general form of MILP problem. \textbf{b,} Three types of NP-Hard MILP problem included in our testing dataset. \textbf{c,} A complex large-scale set covering problem in MILP form. The matrices $A$, $b$, and $c^{T}$ in this example are highlighted in corresponding colors. \textbf{d,} Part of the branch-and-bound tree obtained by solving the set covering problem in \textbf{c}. Path to the optimal solution are marked in light red. Statistical analysis reveals a severe imbalance in different node types.}
\label{fig1-example}
\end{figure}

Recent advancements in large language models (LLMs) have emerged as a powerful tool, which was introduced in works to explore the foundation model of solving complex MILP problems. For example, Li et al. \cite{li2025foundationmodelsmixedinteger} introduced MILP-Evolve to improve the generalization capability of LLM-based method by an evolutionary framework while Ye et al. \cite{ye2025large} introduced LLM-LNS, which promoted solving large-scale MILP problems using LLM-driven large neighborhood search framework combined with existing heuristics. Awasthi et al. \cite{awasthi2025combinatorialoptimizationllmdriveniterated} demonstrated that LLMs can balance the flexibility of locally expressed constraints with rigorous global optimization by introducing iterated fine-tuning framework where algorithmic feedback progressively refines the LLM's output.


However, to our knowledge, only limited studies have investigated the potential of machine learning-enhanced node selection strategies. Among these, He et al. \cite{he2014learning} were pioneers in training support vector machine (SVM)-based node selection and pruning models using imitation learning. Building upon this foundation, Yilmaz et al. \cite{yilmaz2021study} conducted an extensive investigation into the impact of various node feature combinations within a similar experimental framework. Song et al. \cite{song2018learning} learned a node selection by imitation learning of shortest paths to good feasible solutions. While their method achieved scale generalization by incrementally increasing the problem size during training, this approach was computationally intensive and potentially compromised the method's generalizability. 

Labassi et al. \cite{labassi2022learning} used the bipartite graph representation from Gasse et al. \cite{gasse2019exact} to encode the state of nodes. They treated the node selection process as a pairwise comparison between two different nodes, training their model to distinguish nodes that contain the optimal solution. However, their experimental validation was limited to small-scale MILP problems that could be readily solved by open-source solvers. Recently, Mattick \& Mutschler \cite{mattick2023reinforcement} and Zhang et al. \cite{zhang2025learning} employed the representation of whole branch-and-bound tree to encode the state of nodes and applied reinforcement learning to learn the selection of nodes, but the scale of their benchmark MILP instances was also limited. Notably, unlike variable selection, effective node selection in branch-and-bound method requires identifying nodes that contain better feasible solutions to update the global primal bound \cite{achterberg2009scip, gasse2022machine}. Such nodes are exceedingly rare in the search tree, depicted in Figure 1.d, presenting significant challenges of sample imbalance for training robust machine learning models.

\subsubsection*{Designing faster mixed integer linear programming
algorithm via DeepBound}
We propose a novel approach, named DeepBound, to learn to prioritize the optimal path nodes, which contain the optimal solution to the original MILP problem, in the branch-and-bound tree. To tackle the two critical challenges of node feature encoding and data imbalance faced by learning-based methods in node selection. The core design of DeepBound introduces a multi-level node feature fusion network, complemented by a pairwise training mechanism. This architecture enables DeepBound to process paired node feature data generated during each step of the branch-and-bound procedure, performing cross-feature and cross-node dimension information fusion to enhance its ability to identify nodes belonging to the optimal path.

We evaluate the node selection capability of DeepBound on various MILP problem datasets. Integrated with the state-of-the-art SCIP solver \cite{achterberg2009scip}, our experiments compare the acceleration provided by DeepBound against traditional human-designed heuristics and other learning-based methods when solving practical MILP problems (e.g., the set covering problem instance in Figure 1.c). Furthermore, our results demonstrate that the node selection model learned by DeepBound on smaller MILP datasets exhibits strong generalization capabilities, effectively handling more complex problems with increased numbers of variables and constraints.

\section{Results}\label{sec2}

\begin{figure}[!thbp]
\centering
\includegraphics[width=0.9\columnwidth]{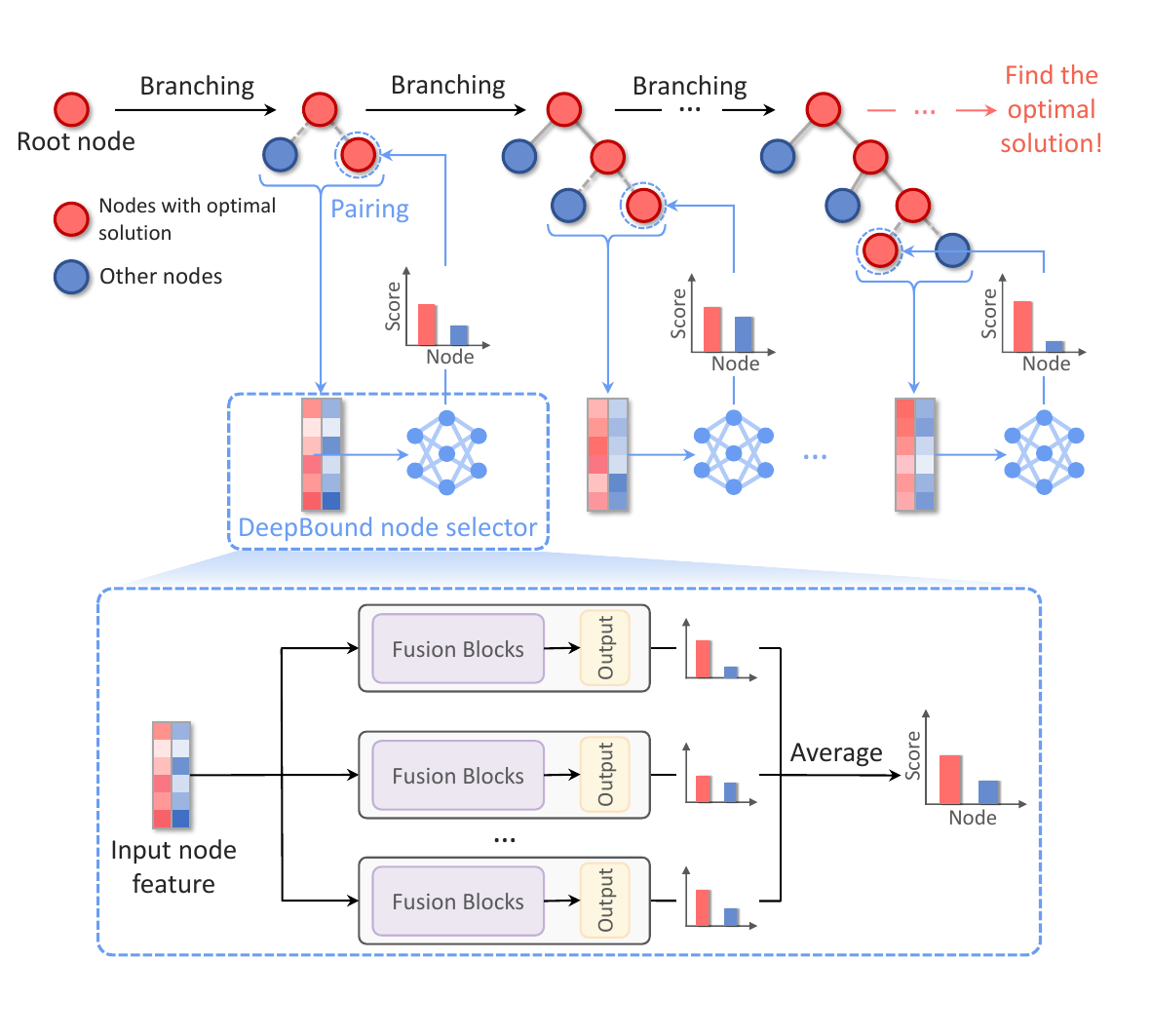}
\caption{\textbf{DeepBound architecture}. Overview of using DeepBound for node selection, sorting nodes based on their scores and iteratively selecting the higher-ranked node each time to find the optimal solution. DeepBound utilizes the ensemble learning mechanism of fusion models to process feature vectors of node pairs and provides average score for node ranking. Each fusion block performs node-level and feature-level fusion on the feature vectors of node pairs for node comparison and subsequent node scoring.}
\label{Ins-overview}
\end{figure}

\subsection{DeepBound learns the optimal solution path}
DeepBound replaces the original heuristic rules in the branch-and-bound algorithm by scoring newly generated node pairs using a neural network based on multi-level feature fusion (Figure 2). In the branch-and-bound algorithm, each newly generated node from a branching operation, along with all unexplored leaf nodes, is placed in a priority queue. The scores assigned to the nodes by the node selection algorithm determine the order of the nodes in the priority queue \cite{achterberg2009scip}. When training our DeepBound network model, we use all nodes along the optimal solution path in the branch-and-bound tree of an already solved MILP problem as \textbf{oracle nodes}, and train the DeepBound model to assign higher scores to these nodes. However, such nodes account for only a small proportion of the branch-and-bound tree, leading to a dramatic imbalance problem.

\subsubsection*{Mitigating node imbalance via pairing node samples}
To address this imbalance issue, we introduce a pairwise training protocol and a learning-to-rank approach. In this protocol, the oracle nodes from the training data are paired with non-oracle nodes from the same priority queue to form training pairs, converting the original objective into learning a node ranking strategy. Specifically, the DeepBound model is trained to assign higher scores to oracle nodes than to non-oracle nodes within the same training pair. By learning this strategy, the oracle nodes rank higher in the priority queue, allowing the MILP solver to select the oracle nodes earlier from the priority queue, positively accelerating the solving process.

\subsubsection*{Enhancing DeepBound via multi-level feature fusion}
In the DeepBound setup, the feature vectors of the two new nodes generated from each new branching operation are input into the model, performing multiple rounds of feature fusion across different dimensions between the nodes and features through a series of concatenated fusion modules (detailed in the Methods, Figure C3). These fusion modules are implemented using different MLPs. Additionally, multiple fusion networks trained through ensemble learning methods are paralleled, providing robust average scores for the node pairs, which will be sent to the priority queue to determine the order of the newly branched node pairs.

\subsection{DeepBound achieves leadership on challenging MILP benchmarks}\label{subsec2}
Our testing dataset includes three NP-hard MILP problem benchmarks: set covering, combinatorial auction, and capacitated facility location. For each type of MILP problem, we generated two sets of testing instances with varying problem sizes: one set with instances of the same size as those in the training set, and another set with instances larger than those in the training set. These different sizes correspond to varying levels of difficulty in solving the problems. Each testing set contains 100 randomly generated instances. Further details of each MILP problem are included in the Methods.

\begin{figure}[!thbp]
  \centering
  \begin{minipage}{\textwidth}
  \centering
  \begin{subfigure}{.33\columnwidth} 
    \includegraphics[width=\linewidth]{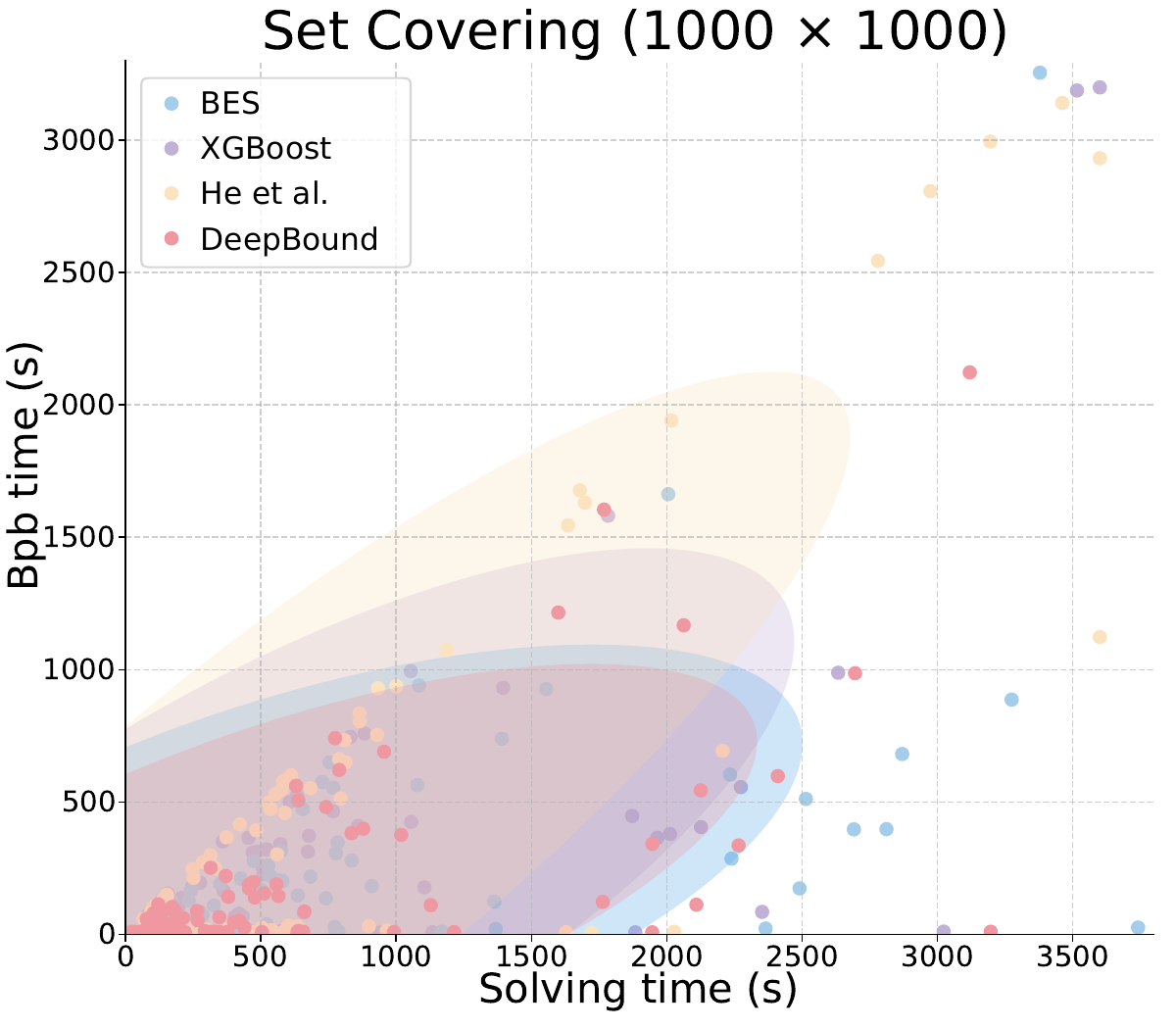}
    \caption{}
    \label{scatter:sub1}
  \end{subfigure}%
  \begin{subfigure}{.33\columnwidth} 
    \includegraphics[width=\linewidth]{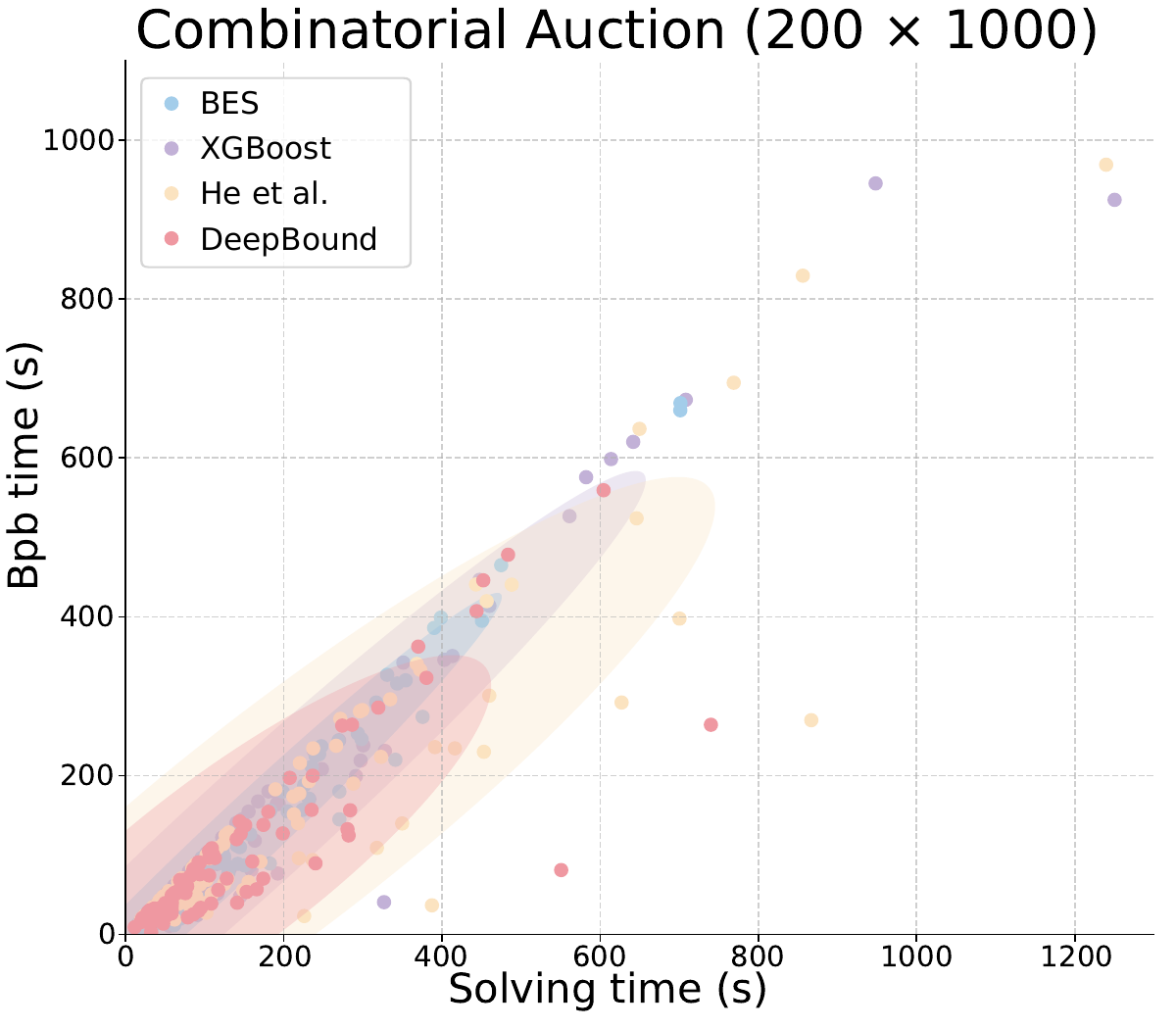}
    \caption{}
    \label{scatter:sub2}
  \end{subfigure}
  \begin{subfigure}{.33\columnwidth} 
    \includegraphics[width=\linewidth]{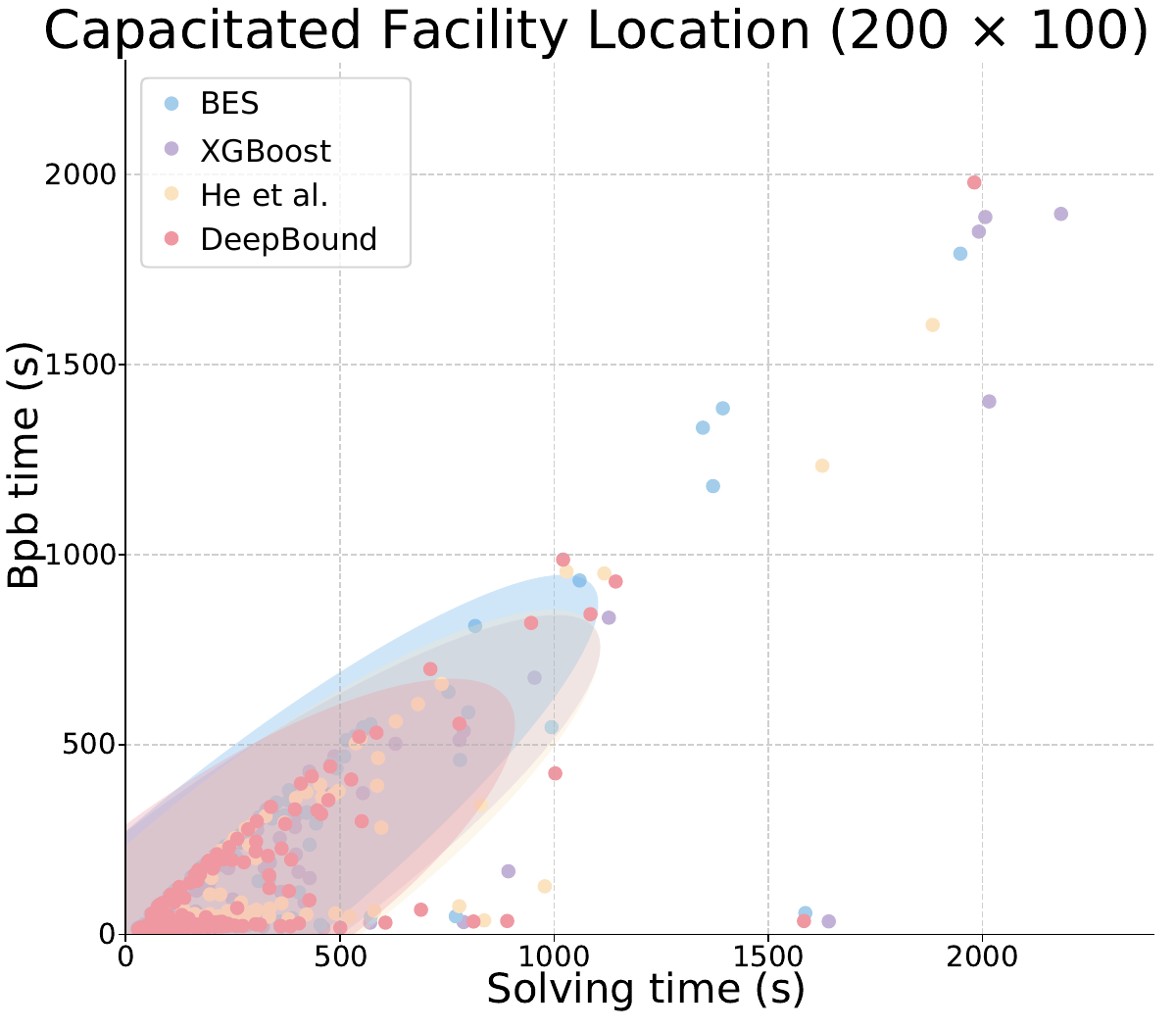}
    \caption{}
    \label{scatter:sub3}
  \end{subfigure}%

  \end{minipage}

  \begin{minipage}{\textwidth}
  \centering
  \begin{subfigure}{.33\columnwidth} 
    \includegraphics[width=\linewidth]{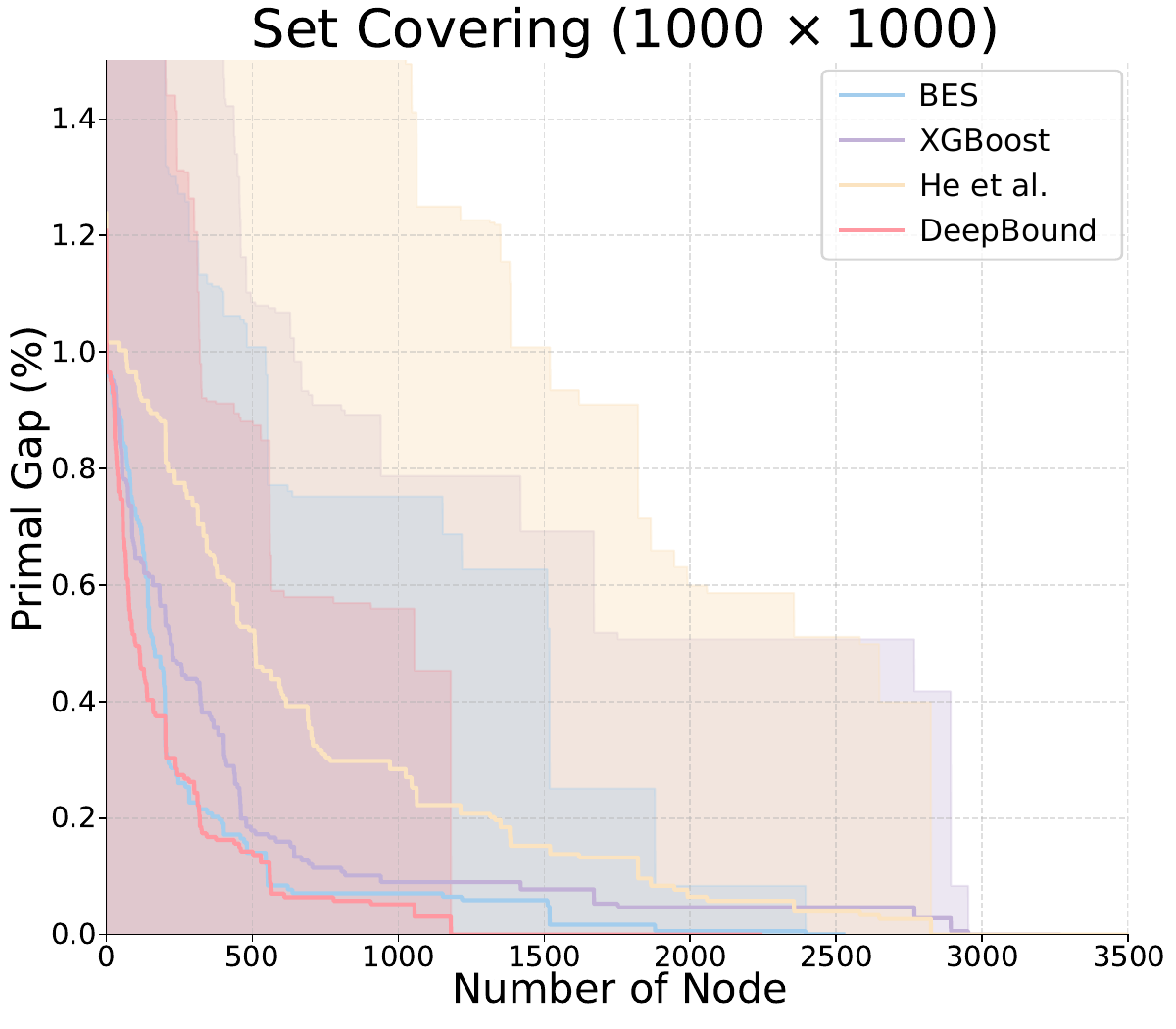}
    \caption{}
    \label{gap:sub1}
  \end{subfigure}%
  \begin{subfigure}{.33\columnwidth} 
    \includegraphics[width=\linewidth]{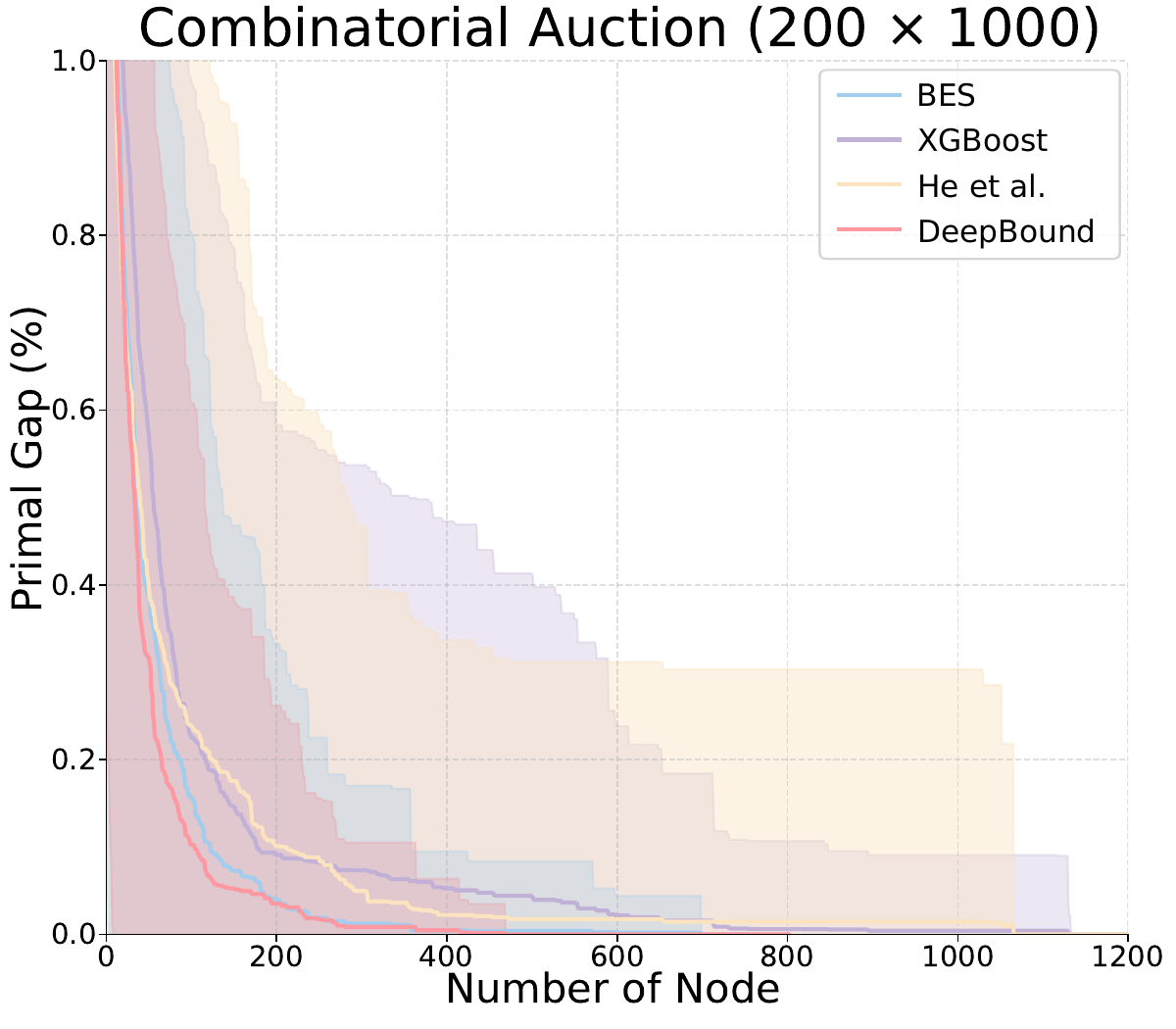}
    \caption{}
    \label{gap:sub2}
  \end{subfigure}
  \begin{subfigure}{.33\columnwidth} 
    \includegraphics[width=\linewidth]{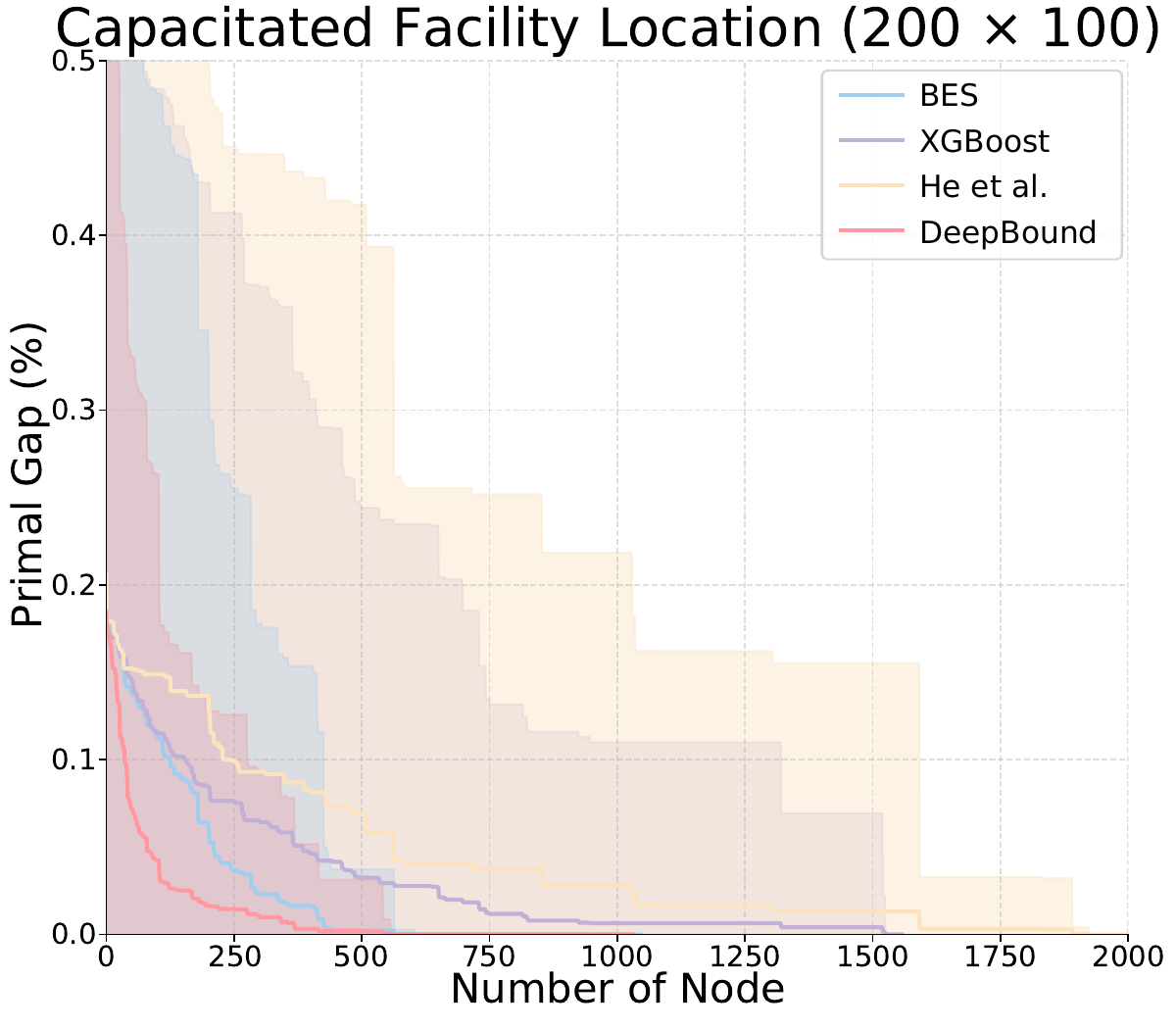}
    \caption{}
    \label{gap:sub3}
  \end{subfigure}%

  \end{minipage}

  \begin{minipage}{\textwidth}
  \centering
  \begin{subfigure}{.33\columnwidth} 
    \includegraphics[width=\linewidth]{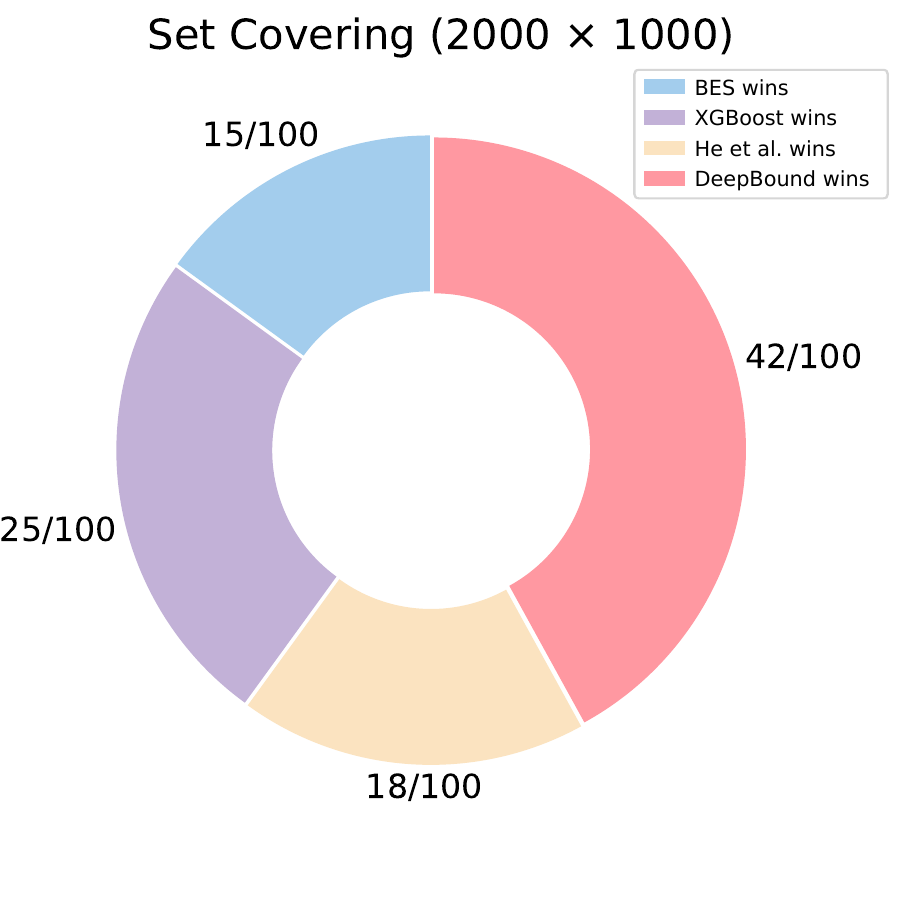}
    \caption{}
    \label{hard:sub1}
  \end{subfigure}%
  \begin{subfigure}{.33\columnwidth} 
    \includegraphics[width=\linewidth]{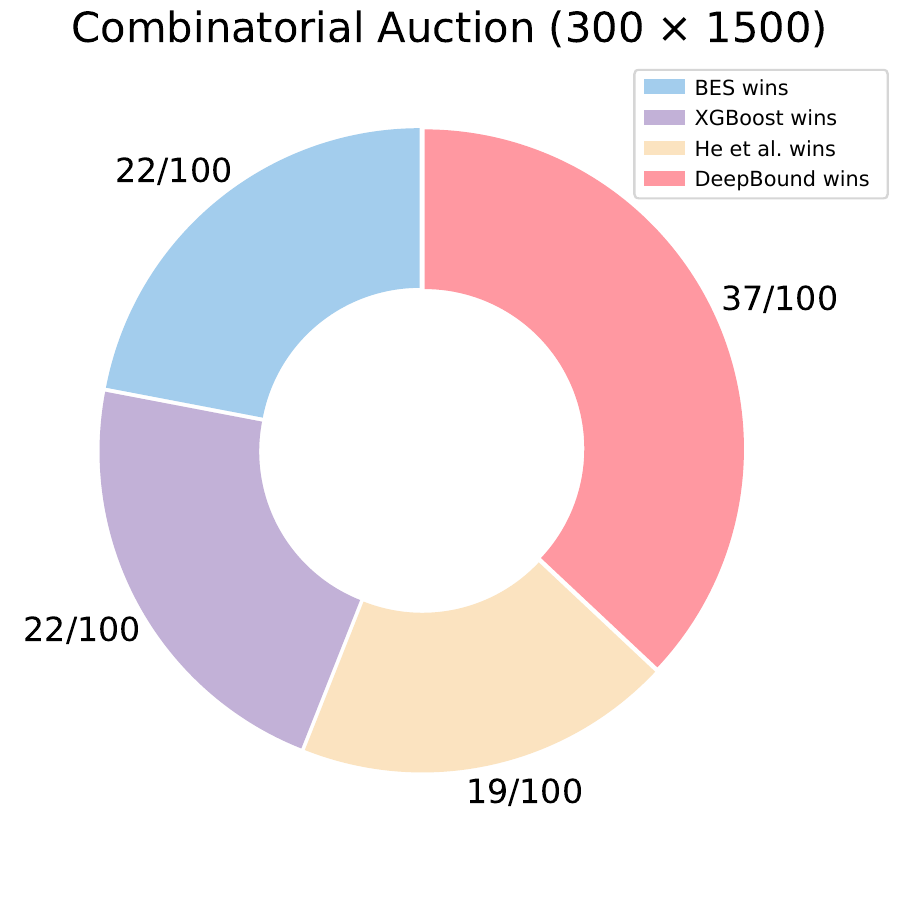}
    \caption{}
    \label{hard:sub2}
  \end{subfigure}
  \begin{subfigure}{.33\columnwidth} 
    \includegraphics[width=\linewidth]{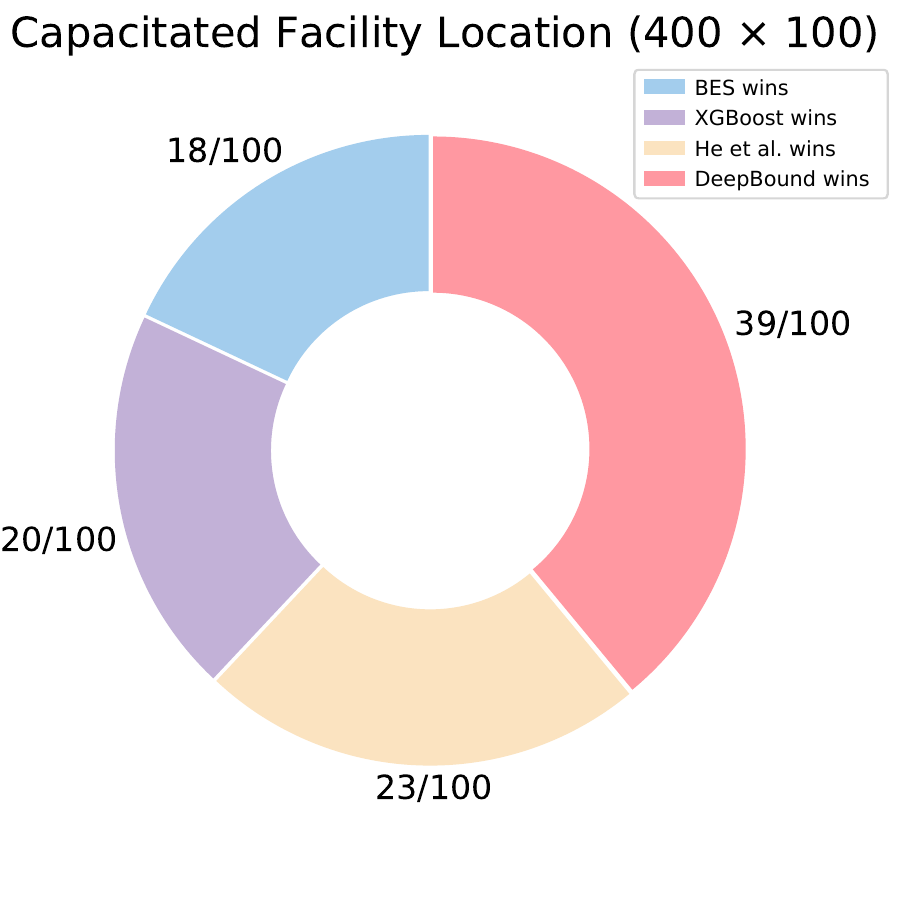}
    \caption{}
    \label{hard:sub3}
  \end{subfigure}%

  \end{minipage}

  \caption{\textbf{Evaluation of DeepBound node selector with MILP test sets of different types and scales.} \textbf{a-c,} Evaluation on instances of three different MILP problems. The figures show the distribution of solving time and bpb time for the different node selection methods on 100 randomly generated instances of each type of MILP problem, with the 95\% confidence ellipses. \textbf{d-f,} Evaluation of the convergence pattern of primal gap for different node selection methods in solving three different MILP problems, with the mean curves and 95\% confidence intervals. \textbf{g-i,} Comparison of the number of different node selection methods that achieve the fastest solving speed (wins) on the same set of 100 large-scale hard problem instances for each type of MILP problem.}
  \label{compare-fig1}
\end{figure}

\begin{figure}[thbp]
\centering
\includegraphics[width=0.9\columnwidth]{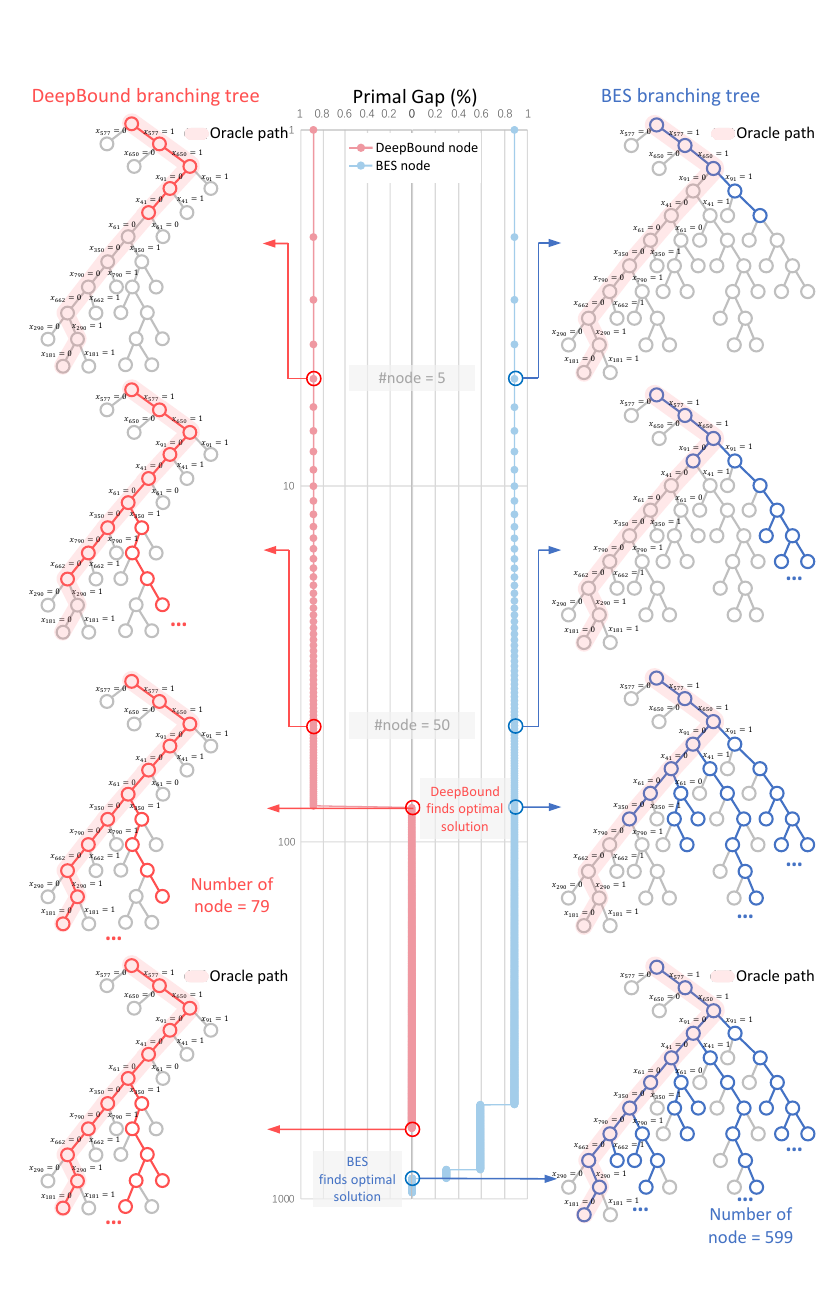}
\caption{\textbf{DeepBound accelerates the solving of MILP problems}. Comparison of the branch-and-bound trees and the node-primal gap curves between DeepBound and BES when solving identical 2000$\times$1000 set covering problems. The central panel illustrates the node-primal gap curves, which depict the relationship between the number of already explored nodes and the primal gap at the node. Figures on either side visualize the branch-and-bound tree of DeepBound and BES at different stages of the solving process. The oracle node paths leading to the discovery of the optimal feasible solution within each tree are emphasized in light red shading.}
\label{Tree-compare}
\end{figure}

We evaluated different node selection algorithms by combining them with the same full-strong branching (FSB) rule used in SCIP \cite{achterberg2009scip}. The total \textbf{solving time} was reported in seconds, including the running time for unsolved instances without additional penalties. Additionally, we incorporated the best primal bound time (\textbf{bpb-time}) to measure the time taken by the node selector to find the optimal feasible solution in the branch-and-bound tree. After this point, the branch-and-bound solver will use this optimal feasible solution for node pruning. Thus, this metric reflects the ability of the node selection algorithm to identify the optimal feasible solution and accelerate the optimization of the global primal bound. 

Our analysis further extended to examining the \textbf{convergence patterns of primal gaps} across different node selection algorithms on the three evaluation datasets. The primal gap, which quantifies the disparity between the objective function value of the integer feasible solution at the current node and the known optimal feasible solution, serves as a crucial metric for evaluating algorithmic performance. The convergence pattern of primal gap provides valuable insights into the efficiency of each node selection algorithm to find the optimal solution.


Finally, in addressing the large-scale, computationally challenging problem sets, not all instances could be completely solved by all selectors within the given computational constraints, we counted the number of instances in which each node selector achieved the fastest solving time (\textbf{wins}) among all four node selection algorithms, thereby providing a quantitative measure of relative performance in solving complex problems.

\subsubsection*{Faster solving of MILP problems using DeepBound}
In terms of solving time and best-primal-bound time (bpb-time), when tested on instances with similar size and complexity to the training set across all three MILP benchmarks, DeepBound consistently outperforms all other node selection methods. Specially, both the solving time and bpb-time distributions of DeepBound are closer to the origin than the competing approaches, including SCIP's default best estimate search (BES) node selection rule \cite{achterberg2009scip} and two machine learning-based methods \cite{he2014learning}. This empirical evidence underscores DeepBound's superior ability to identify optimal solutions more rapidly within the branch-and-bound framework, thereby significantly enhancing the efficiency of exact MILP solutions. The results indicate that DeepBound achieves these improvements without compromising solution quality, demonstrating its effectiveness in optimizing the search process for MILP problems.

The effectiveness of DeepBound in accelerating the finding of nodes containing optimal solutions can be further demonstrated through an evaluation of primal gap metrics. The primal gap represents the discrepancy between the objective function value of the best-so-far integer feasible solution and the optimal solution within the branch-and-bound framework. For each MILP test set, we analyzed the average convergence behavior of primal gaps across different node selection strategies, incorporating the 95\% confidence intervals for all test samples. As illustrated in Figures 3d-f, DeepBound consistently demonstrates superior performance, achieving the fastest primal gap convergence across all three evaluation benchmarks. The algorithm exhibits the most consistent convergence behavior, as evidenced by the smallest confidence interval area across test instances. This robust acceleration in primal gap reduction strongly corroborates the promoting effect of DeepBound in efficiently locating optimal solutions within branch-and-bound method.

\subsubsection*{DeepBound generalizes and scales to larger MILP problems}
In generalization tests, as the problem scale increases, the proportion of problems that can be solved in one hour decreases significantly. Despite the substantial increase in problem scale and difficulty, DeepBound performs well on instances larger than those in the training sets and outperforms both other learning-based node selectors and SCIP's default node selection rule, BES, in terms of the number of wins (Figure 3g-i), which indicates that DeepBound maintains leading performance even facing the more complex MILP problems. 

An example of the acceleration effect of DeepBound in finding the optimal feasible solution is presented in Figure 4. When solving the same instance of a 2000$\times$1000 set covering problem, DeepBound is able to explore the branch containing the optimal feasible solution earlier in the branch-and-bound tree. In contrast, the BES method explores nodes more extensively across the branch-and-bound tree and delves deeper into the branch containing the optimal solution at a later stage. This focused exploration enables DeepBound to find the optimal feasible solution after solving only 79 nodes (where the primal gap drops to 0), whereas the BES algorithm finds the optimal solution after exploring 599 nodes, resulting in a significantly higher total number of nodes compared to DeepBound.

\subsection{DeepBound accelerates selecting the oracle nodes}\label{subsec3}


\begin{figure}[!bp]
  \centering
  
  \begin{subfigure}{.325\columnwidth} 
    \includegraphics[width=\linewidth]{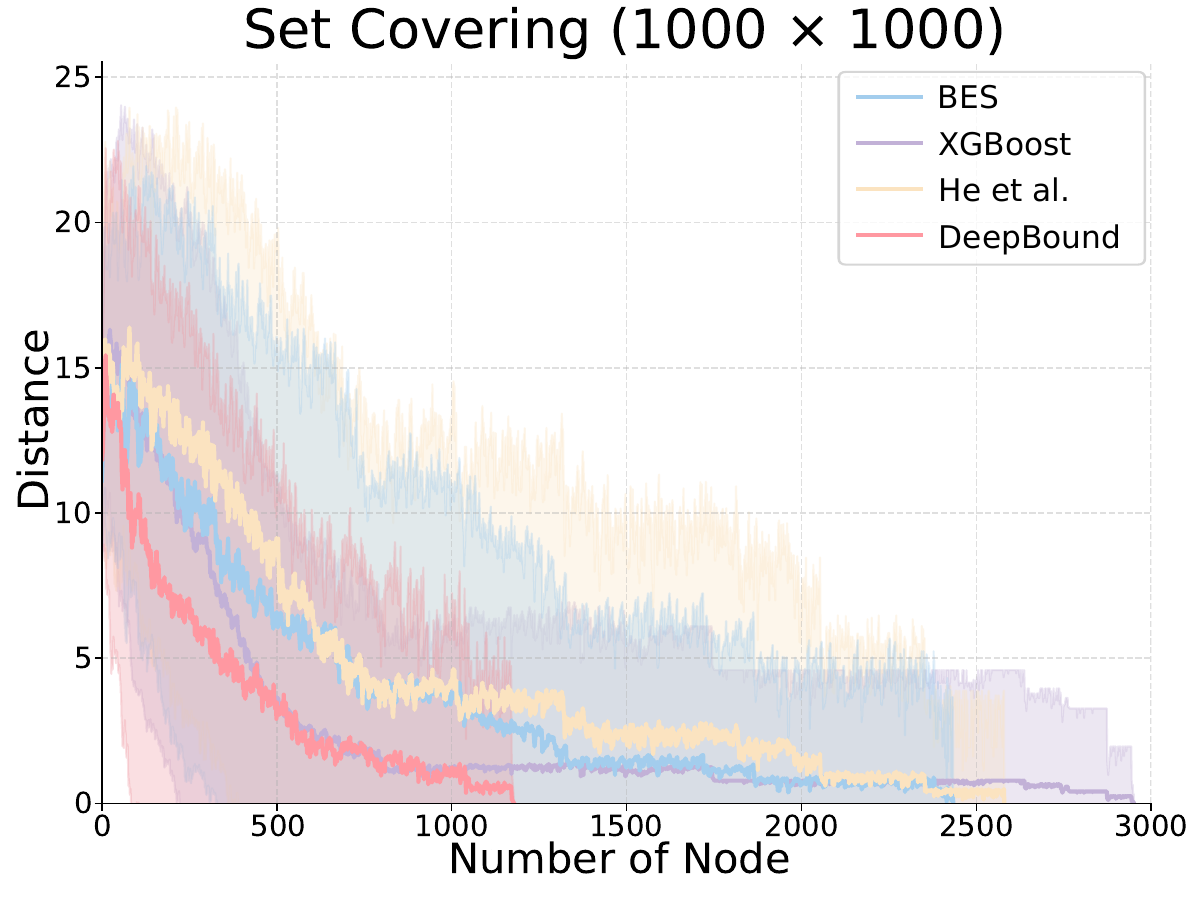}
    \caption{}
    \label{distance:sub1}
  \end{subfigure}%
  \begin{subfigure}{.325\columnwidth} 
    \includegraphics[width=\linewidth]{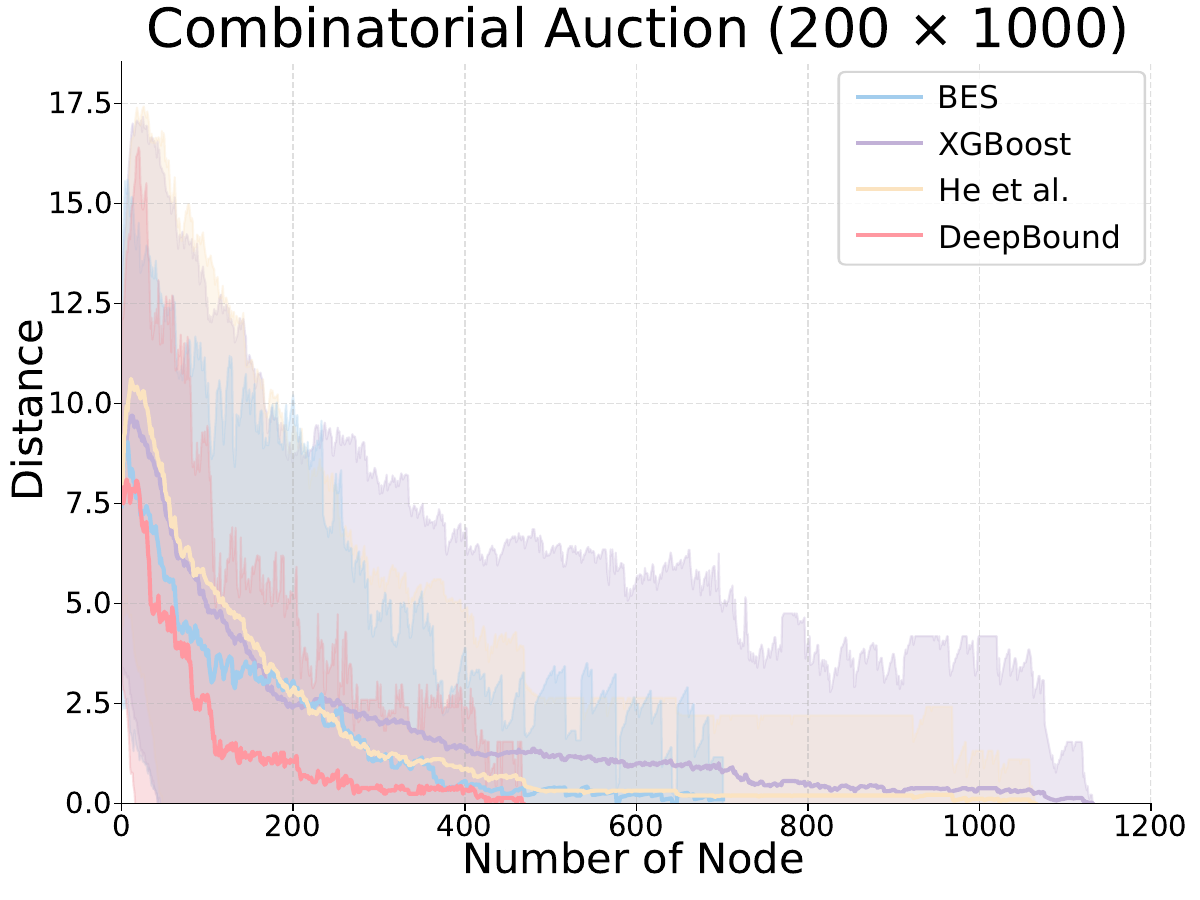}
    \caption{}
    \label{distance:sub2}
  \end{subfigure}
  \begin{subfigure}{.325\columnwidth} 
    \includegraphics[width=\linewidth]{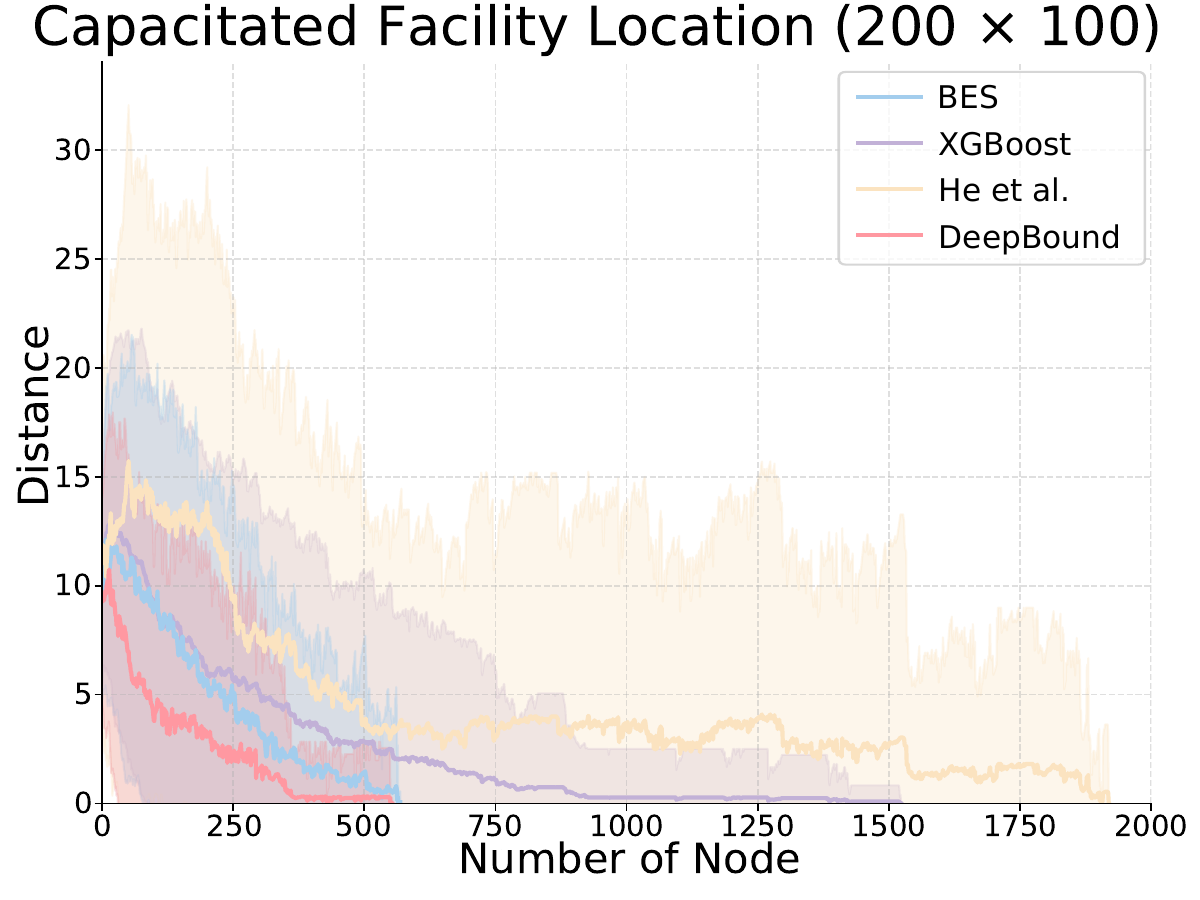}
    \caption{}
    \label{distance:sub3}
  \end{subfigure}%

  \caption{\textbf{Evaluation of the distance to the optimal solution.} \textbf{a-c,} Evaluation of the decreasing trend of the distance for different node selection methods between the current node and the node containing the optimal solution in the branch-and-bound tree before the optimal solution was found. The mean curves and 95\% confidence intervals are depicted.}
  \label{distance}
\end{figure}

To further analyze how the DeepBound algorithm accelerates the branch-and-bound method in finding the optimal feasible solution, we conducted more detailed experiments on the three MILP problems. To quantify the deviation of the search process from the optimal solution path, we define the distance $D(n_i, n_{opt})$ from the current node $n_i$ to the node with optimal solution $n_{opt}$:
\begin{equation}
D(n_i, n_{opt}) = len(his(n_{opt})) + diff(his(n_i),his(n_{opt})) - LCS(his(n_i),his(n_{opt})) .
\end{equation}
where $his(\cdot)$ represents the branching history of a node. The term $diff(his(n_i),his(n_{opt}))$ denotes the number of variables with differing values in the branching histories of two nodes, indicating the number of incorrect node selections. Conversely, $LCS(his(n_i),his(n_{opt}))$ represents the length of the longest common substring between the branching history of two nodes, indicating the number of correct node selections. Incorrect variable values resulting from node selection can cause the current node's solution space to deviate from the optimal solution space, thereby increasing the solving time by creating greater challenges for primal heuristics in the branch-and-bound solver \cite{berthold2006primal}. 

We performed this optimal solution distance analysis on three MILP benchmark problems: the 1000$\times$1000 set covering problem, the 200$\times$1000 combinatorial auction problem, and the 200$\times$100 capacitated facility location problem. The experimental results were compared between the DeepBound algorithm and three other approaches: the default node selection algorithm BES in the SCIP solver, as well as two other machine learning-based algorithms.

\subsubsection*{DeepBound approaches the optimal solution faster}
During the branch-and-bound process, the node distances for all methods exhibit a decreasing trend before the optimal solution is identified. Notably, the DeepBound algorithm demonstrates the most rapid decline in node distance across all three MILP problems (Figure 5a-c), suggesting that the feasible regions of nodes selected by DeepBound are closer to the region containing the optimal solution. This allows DeepBound to identify the optimal solution more quickly within the branch-and-bound tree compared to other methods. Similar acceleration in selecting the oracle node is also observed in Figure 4, further supporting the efficiency of this approach. In contrast, machine learning-based and heuristic algorithms display inconsistent performance across different problems, with no consistent superiority observed.

\begin{figure}[!thbp]
  \centering
  
  \begin{subfigure}{.5\columnwidth} 
    \includegraphics[width=\linewidth]{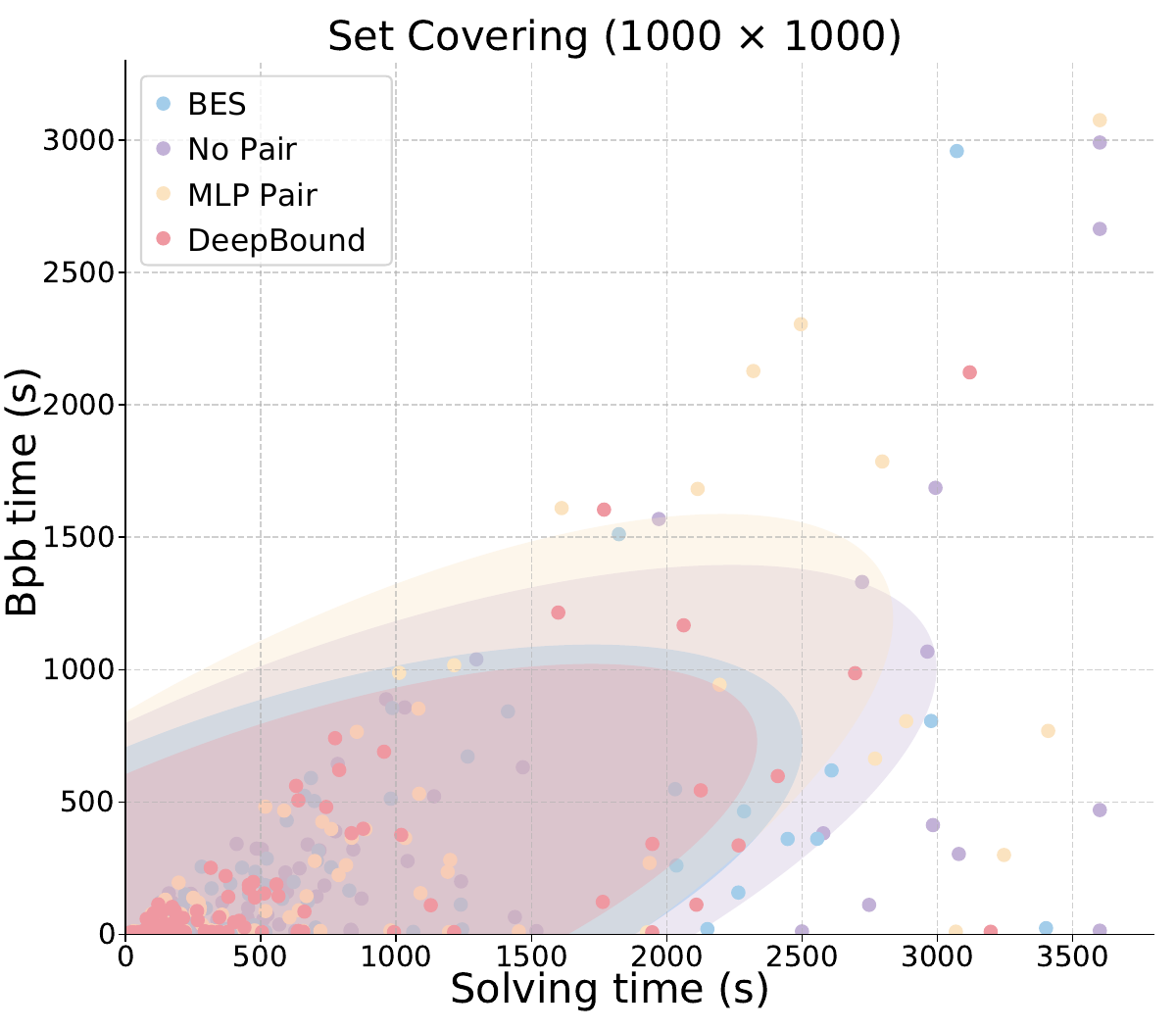}
    \caption{}
    \label{ablation:sub1}
  \end{subfigure}%
  \begin{subfigure}{.5\columnwidth} 
    \includegraphics[width=\linewidth]{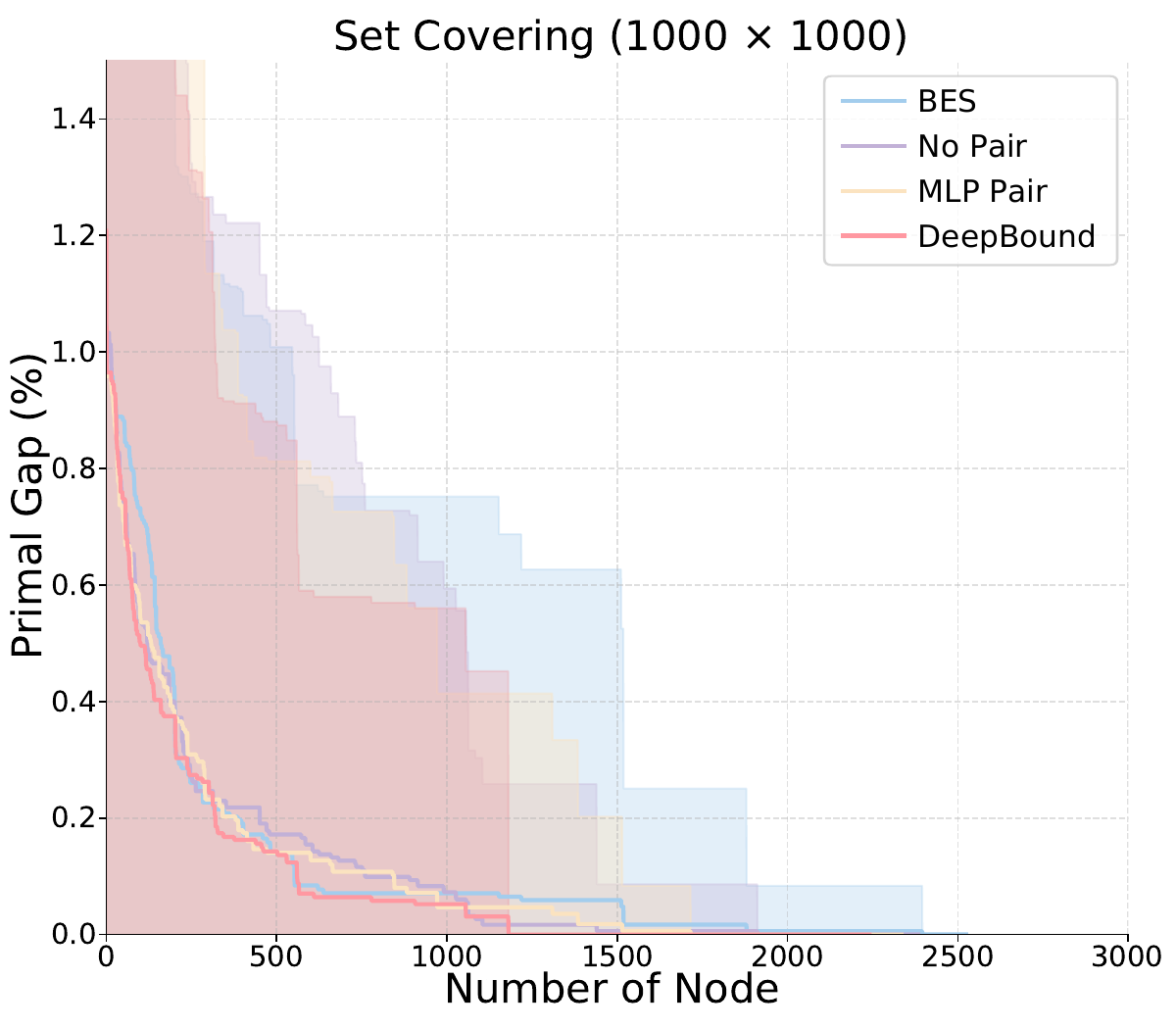}
    \caption{}
    \label{ablation:sub2}
  \end{subfigure}

\caption{\textbf{Evaluation of ablation models of DeepBound.} \textbf{a,} Solving time and bpb time for the
DeepBound, BES, DeepBound with single node feature (\textbf{No Pair}) and DeepBound with vanilla MLP (\textbf{MLP Pair}) on 100 randomly generated instances of set covering 1000$\times$1000 problem, with the 95\% confidence ellipses. \textbf{b,} The comparison of the convergence trend of primal gaps of different node selection methods in solving set covering 1000$\times$1000 problem.}
  \label{of_distribution}
\end{figure}

\subsubsection*{Ablation studies on different model components}
To thoroughly analyze the contributions of different components within the DeepBound framework, we conducted extensive ablation studies. All experiments were performed using the same 1000×1000 set covering problem instances as the test set. We configured an unchanged DeepBound network that only takes single-node features as input (denoted as \textbf{No Pair}) and a MLP network that takes paired node features as input but omits the multi-level feature fusion mechanism (denoted as \textbf{MLP Pair}). This allowed us to isolate and assess the individual impacts of paired feature input and the feature fusion module.

As shown in Figure 6.a, the DeepBound model achieves superior performance, as evidenced by its distributions of solving time and bpb-time being closer to the origin, indicating faster identification of optimal solutions and more efficient exact solving. The primal gap curves in Figure 6.b further confirm this advantage, with DeepBound showing the fastest convergence to zero and maintaining the lowest gap throughout. In contrast, both ablation models and external methods like BES demonstrate inferior performance, and BES additionally suffers from delayed convergence due to its slower discovery of optimal solutions in certain problem instances.

\subsection{Feature analysis of DeepBound}\label{subsec4}                                                                 
\begin{figure}[thbp]
\centering
\includegraphics[width=0.85\columnwidth]{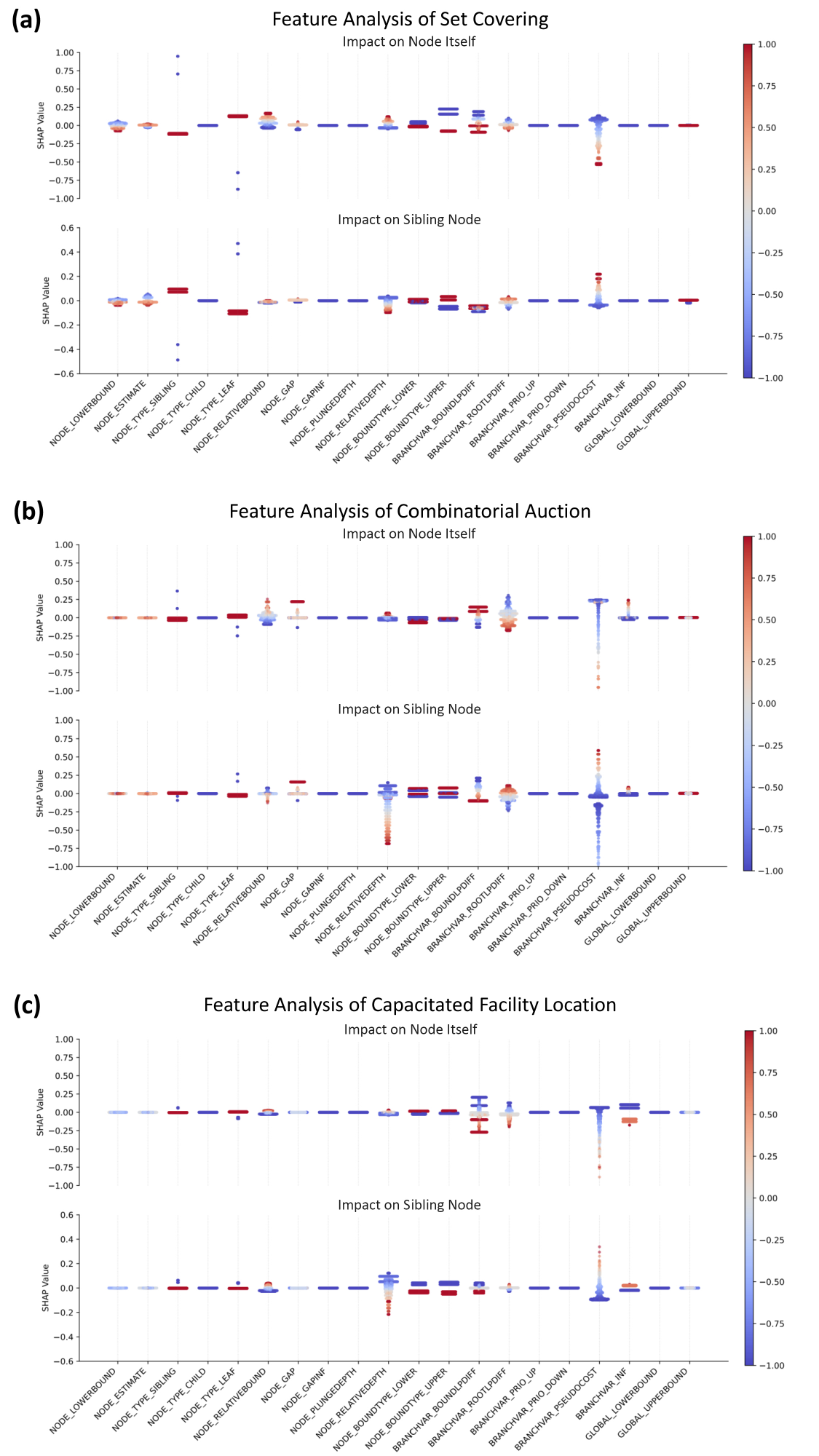}
\caption{\textbf{Feature analysis of DeepBound on different MILP problems.} The SHAP values of all input features on \textbf{a,} set covering, \textbf{b,} combinatorial auction and \textbf{c,} capacitated facility location problem. The color of the data points represents the normalized magnitude of each feature's SHAP value. The impact of input features on the node itself (first row) and on its paired sibling node (second row) are separately analyzed for each problem.}
\label{feature_analysis}
\end{figure}

We conducted a comprehensive analysis of how different input features contribute to the final node scores, aiming to gain deeper insights into the feature relationships that the DeepBound model learns during node selection. Unlike many existing neural network-based methods for solving MILP problems, which typically rely on direct encoding of raw MILP formulations, we adopt a fundamentally different approach. Specifically, DeepBound leverages a more comprehensive set of engineered features to represent nodes and employs neural networks to learn and score within this feature space. 

\subsubsection*{Informative features reduce complexity and enable analysis}
Our input features are meticulously designed to capture three key aspects: node-specific attributes, branching variable information, and global characteristics of the branch-and-bound tree (as detailed in the Methods section). This approach provides two significant advantages: first, by constructing a compact and meaningful feature representation, we substantially reduce the complexity and dimensionality of node descriptions, thereby lowering the learning difficulty for the network; second, our feature representation is consistent with the node selection criteria used in traditional MILP solvers, which facilitates both theoretical comparison and empirical analysis of feature importance across diverse MILP problems. These advantages allow us to not only better understand the underlying mechanisms of the model but also to identify which features are most influential in determining node scores.

We employed the DeepExplainer from the SHAP (SHapley Additive exPlanations) framework to systematically analyze and visualize the influence of input features on the output of the DeepBound model. The SHAP value of each point indicates whether the feature promotes the model's classification of a node as oracle node (positive SHAP value) or non-oracle node (negative SHAP value). The absolute value of the SHAP value reflects the strength of this promotion. The feature analysis for the set covering problem, the combinatorial auction problem and the capacitated facility location problem are presented in Figure 7.a, 7.b, and 7.c, respectively. 

Additionally, we list the features associated with three commonly used node selection rules in SCIP in Table 1. Given that the DeepBound model operates on paired-node features as input and generates scores for each node in the pair, for each problem in Figure 7, we analyzed the impact of node features on two aspects: the node itself (depicted in the first row of each figure) and its paired sibling node (shown in the second row). 

\begin{table}[!h]
\caption{Features used by different node selection heuristic rules.}
\label{heur-feature}

\begin{tabular}{@{}ll@{}}
\toprule
\textbf{Node selection rule} & \textbf{Features} \\
\midrule
Depth First Search    & Node\_Depth  \\
\midrule
\multirow{2}{*}{Best First Search} 
& Node\_Lowerbound  \\
& Node\_Type (with plunging)  \\
\midrule
\multirow{5}{*}{Best Estimate Search}   
& Node\_Lowerbound \\  
& Node\_Estimate  \\
& Node\_Type (with plunging)  \\
& BranchVar\_BoundLPDiff \\ 
& BranchVar\_Pseudocost  \\
\botrule
\end{tabular}

\end{table}

\subsubsection*{DeepBound learns problem-specific feature patterns}
The experimental results reveal a significant mutual influence between the features of the paired nodes, as evidenced by the strong response of the scoring results to the features of the sibling nodes in the second row. These findings indicate that the paired-node input mechanism employed by the DeepBound method, combined with the feature mixing network architecture, provides a more informative and comprehensive representation compared to single-node input approaches.

Manually designed heuristic rules often face the challenge of limited applicability across diverse MILP problems. Unlike these heuristic node selection rules, which depend on fixed feature sets, the DeepBound model can be trained on problem-specific datasets to identify effective feature combinations tailored to each particular problem structure. For the set covering problem, the model's predictions are significantly influenced by node's lower bound, estimate, type, the deviation of the branching variable from integer values, and pseudocosts of branching variable. While these features partially overlap with those used in BFS and BES rules, but the node depth, which is a focus of DFS, shows minimal impact on the model's performance. In contrast, for the combinatorial auction problem, the importance of the type of node decreases, while node depth becomes a much more significant factor. For the capacitated facility location problem, the contribution of node type is minimal, and the impact of the node's lowerbound and estimate on the node score is also negligible. 

This cross-problem analysis reveals that existing heuristic rules fail to adequately account for the diverse feature importance across different MILP problem types. By leveraging extensive node data from various problem instances during training, the DeepBound model can identify feature combinations that are particularly effective in solving specific problems. This capability provides valuable insights for developing simpler yet highly effective heuristic rules that can adapt to different problem structures.

\section{Discussion}\label{sec3}

DeepBound concentrates on learning and replacing the heuristic rules employed during the node selection phase of the branch-and-bound algorithm, instead of attempting to train an end-to-end model of the entire branch-and-bound process. This approach not only mitigates the learning complexity of the neural network model, but also effectively accelerates the solving process when integrated with existing branch-and-bound solvers. Unlike conventional neural network-based methods, DeepBound eschews the direct encoding of raw node information and instead employs node features that are collected in a manner akin to heuristic rules yet with a substantially broader scope for feature selection and collection. Our feature analysis reveals that, unlike traditional heuristic rules, DeepBound calculates these features during node selection via neural network inference. By successfully identifying key features while avoiding the limitations of hard-coded feature calculations, DeepBound offers valuable insights for designing new heuristic algorithm.

Prior research has explored the application of deep learning techniques to substitute specific heuristic rules within the branch-and-bound algorithm. However, many of these approaches have focused exclusively on operations within a single node, neglecting the interdependency between nodes on a larger scale. In contrast, DeepBound employs a multi-level feature fusion network to comprehensively capture both intra-node features and inter-node relationships from the nodes' data. This enhanced information capture is crucial for effective decision-making during the branch-and-bound solving process. Consequently, DeepBound is better equipped to evaluate and encode whether a node contains the optimal solution, which enhances the improvement of node selection methods and expands the scope of data-driven substitution of heuristic rules within the branch-and-bound framework.

Recent advancements in algorithm design leveraging large language models (LLMs) have emerged as a powerful tool, particularly in addressing complex problems such as the cap set problem, where they have surpassed traditional heuristic approaches. Notably, the successful algorithms developed by LLMs often rely on relatively straightforward Python code or pseudocode implementations. In contrast, exact algorithms for mixed-integer programming (MIP) problems, such as branch-and-bound methods, are inherently complex and challenging to express in simple code or natural language. This discrepancy presents a promising opportunity to integrate neural networks, trained on problem-solving data, to enhance or replace heuristic search components within the branch-and-bound framework. 

In the future, a particularly promising research direction is to harness the encoding and representation capabilities of LLMs to either substitute heuristic components or augment exact solving algorithms for MIP problems. By incorporating advanced techniques such as reinforcement learning and chain-of-thought, LLMs could be empowered to perform more efficient reasoning and exploration, thereby significantly improving the performance of MIP solvers.

\bibliography{sn-bibliography}

\newpage
\begin{appendices}

\section{Extended experimental results}\label{secA1}

The following pages present more detailed results of various methods in solving different MILP instances. All experiments were conducted under identical hardware configurations with Intel(R) Xeon(R) Gold 6238R CPU @ 2.20GHz, except that DeepBound utilized the same CPU with a single NVIDIA A100 (80GB) GPU.

\sisetup{detect-weight=true,detect-inline-weight=math,detect-mode=true} 
\begin{table}[!h]
\captionsetup{justification=justified} 
\caption{Policy evaluation on separate instances in terms of number of branch-and-bound nodes, best primalbound time, solving time, and number of wins (fastest method) over number of solved hard instances. For each type of MILP problem, the models are trained on easy instances only. See Section 2.1 for definitions.}
\label{table:results-test}
\centering
\setlength{\tabcolsep}{4.5pt}
\aboverulesep = 0.1mm  
\belowrulesep = 0.2mm  
\begin{tabular}{
    c
    S[table-format=2.2]@{}
    S[table-format=2.2]@{\%\,\,}
    S[table-format=2.2]@{\:\( \pm \)}
    S[table-format=3.2]@{\%\,\,}
    S[table-format=3.0]@{\:\( \pm \)}
    S[table-format=2.1]@{\%\,\,}
    S[table-format=4.2]@{\:\( \pm \)}
    S[table-format=3.1]@{\%\,\,}
    S[table-format=3.0]@{\:/\,}
    S[table-format=3.0]@{\,\,}
    S[table-format=4.0]@{\:\( \pm \)}
    S[table-format=4.]@{\%\,\,}
    S[table-format=3.2]@{\:\( \pm \)}
    S[table-format=2.5]@{\%\,\,}
    S[table-format=2.0]@{\:/\,}
    S[table-format=3.0]@{\,\,}
}

     &
    \multicolumn{6}{ c }{\textbf{Easy}} &
    \multicolumn{6}{ c }{\textbf{Medium}} &
    \multicolumn{4}{ c }{\textbf{Hard}} \\

    Method &
    \multicolumn{2}{ c }{Nodes} &
    \multicolumn{2}{ c }{Bpb-time} &
    \multicolumn{2}{ c }{Sol-time} &
    \multicolumn{2}{ c }{Nodes} &
    \multicolumn{2}{ c }{Bpb-time} &
    \multicolumn{2}{ c }{Sol-time} &
    \multicolumn{2}{ c }{Nodes} &
    \multicolumn{2}{ c }{Wins} \\

    \toprule
    

    \textsc{scip-bes}
        & \multicolumn{2}{ c }{23} & \multicolumn{2}{ c }{15.40} & \multicolumn{2}{ c }{22.34}
        & \multicolumn{2}{ c }{418} & \multicolumn{2}{ c }{224.12} & \multicolumn{2}{ c }{394.83}
        & \multicolumn{2}{ c }{463} & 15 & 100 \\

    \cmidrule(lr){1-7} \cmidrule(lr){8-13} \cmidrule(lr){14-17}

    XGBoost
        & \multicolumn{2}{ c }{24} & \multicolumn{2}{ c }{14.75} & \multicolumn{2}{ c }{21.36}
        & \multicolumn{2}{ c }{395} & \multicolumn{2}{ c }{147.37} & \multicolumn{2}{ c }{383.46}
        & \multicolumn{2}{ c }{455} & 25 & 100 \\
        
    He et al.\cite{he2014learning}
        & \multicolumn{2}{ c }{24} & \multicolumn{2}{ c }{15.47} & \multicolumn{2}{ c }{21.98}
        & \multicolumn{2}{ c }{399} & \multicolumn{2}{ c }{158.29} & \multicolumn{2}{ c }{398.71}
        & \multicolumn{2}{ c }{483} & 18 & 100 \\

    DeepBound (Ours)
        & \multicolumn{2}{ c }{23} & \multicolumn{2}{ c }{\textbf{14.41}} & \multicolumn{2}{ c }{\textbf{21.23}}
        & \multicolumn{2}{ c }{\textbf{370}} & \multicolumn{2}{ c }{\textbf{118.94}} & \multicolumn{2}{ c }{\textbf{371.39}}
        & \multicolumn{2}{ c }{\textbf{426}} & \textbf{42} & \textbf{100} \\

    \cmidrule(lr){1-7} \cmidrule(lr){8-13} \cmidrule(lr){14-17} 
    \\[-8pt]
    \multicolumn{7}{ c }{1000$\times$500} & 
    \multicolumn{6}{ c }{1000$\times$1000} &
    \multicolumn{4}{ c }{1000$\times$2000} 
    \\[-1pt]
    & \multicolumn{12}{ c }{Set Covering} \\
    \\[-3pt]
    \cmidrule(lr){1-7} \cmidrule(lr){8-13} \cmidrule(lr){14-17}

    \textsc{scip-bes}
        & \multicolumn{2}{ c }{13} & \multicolumn{2}{ c }{5.91} & \multicolumn{2}{ c }{6.74}
        & \multicolumn{2}{ c }{88} & \multicolumn{2}{ c }{67.82} & \multicolumn{2}{ c }{93.48}
        & \multicolumn{2}{ c }{431} & 22 & 100 \\

    \cmidrule(lr){1-7} \cmidrule(lr){8-13} \cmidrule(lr){14-17}

    \textsc{XGBoost}
        & \multicolumn{2}{ c }{13} & \multicolumn{2}{ c }{5.84} & \multicolumn{2}{ c }{7.12}
        & \multicolumn{2}{ c }{100} & \multicolumn{2}{ c }{72.56} & \multicolumn{2}{ c }{101.44}
        & \multicolumn{2}{ c }{451} & 22 & 100 \\
        
    He et al.\cite{he2014learning}
        & \multicolumn{2}{ c }{13} & \multicolumn{2}{ c }{5.64} & \multicolumn{2}{ c }{6.90}
        & \multicolumn{2}{ c }{98} & \multicolumn{2}{ c }{69.04} & \multicolumn{2}{ c }{99.60}
        & \multicolumn{2}{ c }{418} & 19 & 100 \\

    DeepBound (Ours)
        & \multicolumn{2}{ c }{13} & \multicolumn{2}{ c }{\textbf{5.42}} & \multicolumn{2}{ c }{\textbf{6.67}}
        & \multicolumn{2}{ c }{\textbf{78}} & \multicolumn{2}{ c }{\textbf{61.16}} & \multicolumn{2}{ c }{\textbf{88.52}}
        & \multicolumn{2}{ c }{\textbf{310}} & \textbf{37} & \textbf{100} \\

    \cmidrule(lr){1-7} \cmidrule(lr){8-13} \cmidrule(lr){14-17}

    \\[-8pt]
    \multicolumn{7}{ c }{100$\times$500} & 
    \multicolumn{6}{ c }{200$\times$1000} &
    \multicolumn{4}{ c }{300$\times$1500} 
    \\[-1pt]
    & \multicolumn{12}{ c }{Combinatorial Auction} \\
    \\[-3pt]
    \cmidrule(lr){1-7} \cmidrule(lr){8-13} \cmidrule(lr){14-17}

    \textsc{scip-bes}
        & \multicolumn{2}{ c }{56} & \multicolumn{2}{ c }{14.40} & \multicolumn{2}{ c }{36.45}
        & \multicolumn{2}{ c }{76} & \multicolumn{2}{ c }{59.30} & \multicolumn{2}{ c }{168.02}
        & \multicolumn{2}{ c }{74} & 18 & 100 \\

    \cmidrule(lr){1-7} \cmidrule(lr){8-13} \cmidrule(lr){14-17}

    \textsc{XGBoost}
        & \multicolumn{2}{ c }{61} & \multicolumn{2}{ c }{14.00} & \multicolumn{2}{ c }{37.48}
        & \multicolumn{2}{ c }{81} & \multicolumn{2}{ c }{55.93} & \multicolumn{2}{ c }{164.97}
        & \multicolumn{2}{ c }{85} & 20 & 100 \\
        
    He et al.\cite{he2014learning}
        & \multicolumn{2}{ c }{63} & \multicolumn{2}{ c }{14.21} & \multicolumn{2}{ c }{37.13}
        & \multicolumn{2}{ c }{83} & \multicolumn{2}{ c }{57.19} & \multicolumn{2}{ c }{169.33}
        & \multicolumn{2}{ c }{92} & 23 & 100 \\

    DeepBound (Ours)
        & \multicolumn{2}{ c }{\textbf{55}} & \multicolumn{2}{ c }{\textbf{13.78}} & \multicolumn{2}{ c }{\textbf{36.10}}
        & \multicolumn{2}{ c }{\textbf{67}} & \multicolumn{2}{ c }{\textbf{52.24}} & \multicolumn{2}{ c }{\textbf{157.56}}
        & \multicolumn{2}{ c }{\textbf{68}} & \textbf{39}  & \textbf{100} \\

    \cmidrule(lr){1-7} \cmidrule(lr){8-13} \cmidrule(lr){14-17}

    \\[-8pt]
    \multicolumn{7}{ c }{100$\times$100} & 
    \multicolumn{6}{ c }{200$\times$100} &
    \multicolumn{4}{ c }{400$\times$100} 
    \\[-1pt]
    & \multicolumn{12}{ c }{Capacitated Facility Location} \\
    \\[-3pt]
    
\end{tabular}
\end{table}

\begin{figure}[thbp]
\centering
\includegraphics[width=0.9\columnwidth]{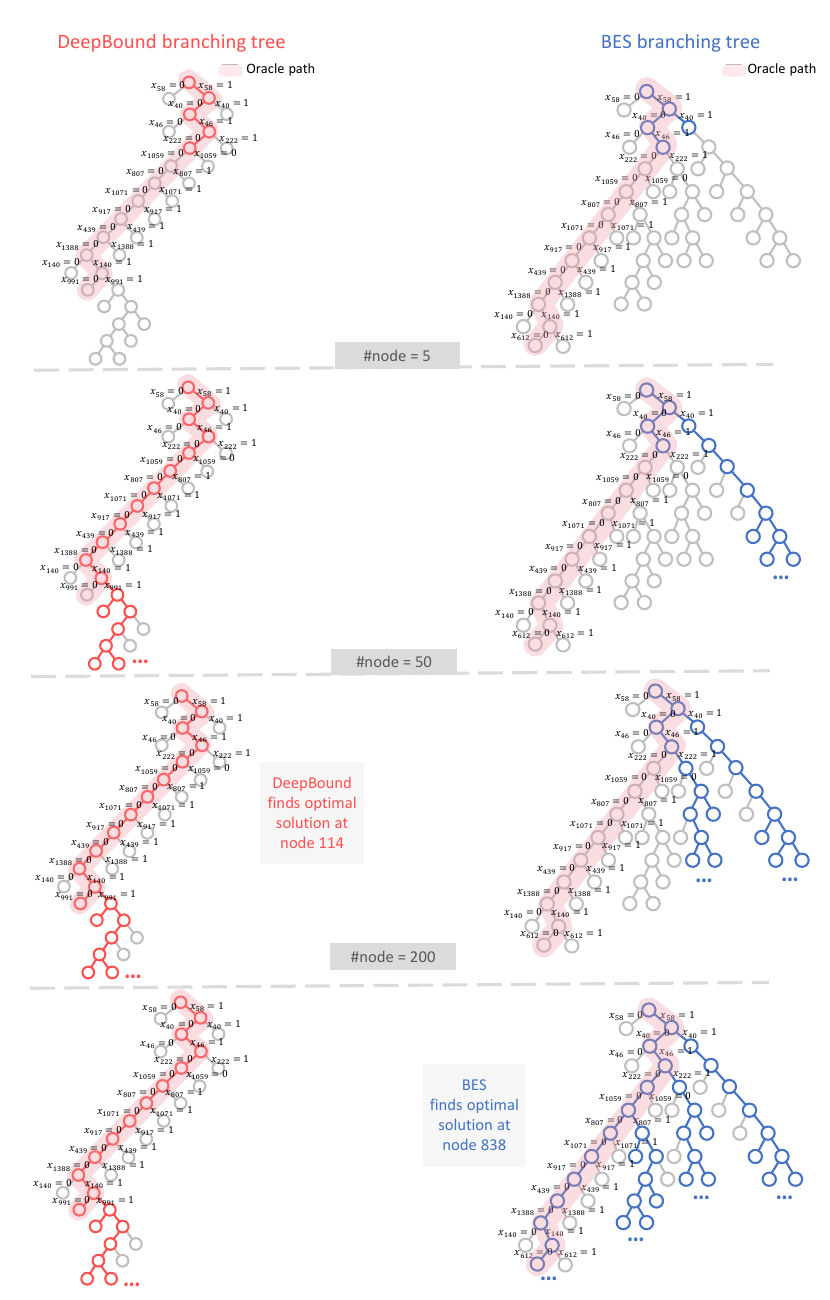}
\caption{\textbf{DeepBound accelerates the solving of MILP problems}. Comparison of the branch-and-bound trees between DeepBound and BES when solving identical 300$\times$1500 combinatorial auction problem. 
}
\label{Tree-compare}
\end{figure}

\begin{figure}[thbp]
\centering
\includegraphics[width=0.9\columnwidth]{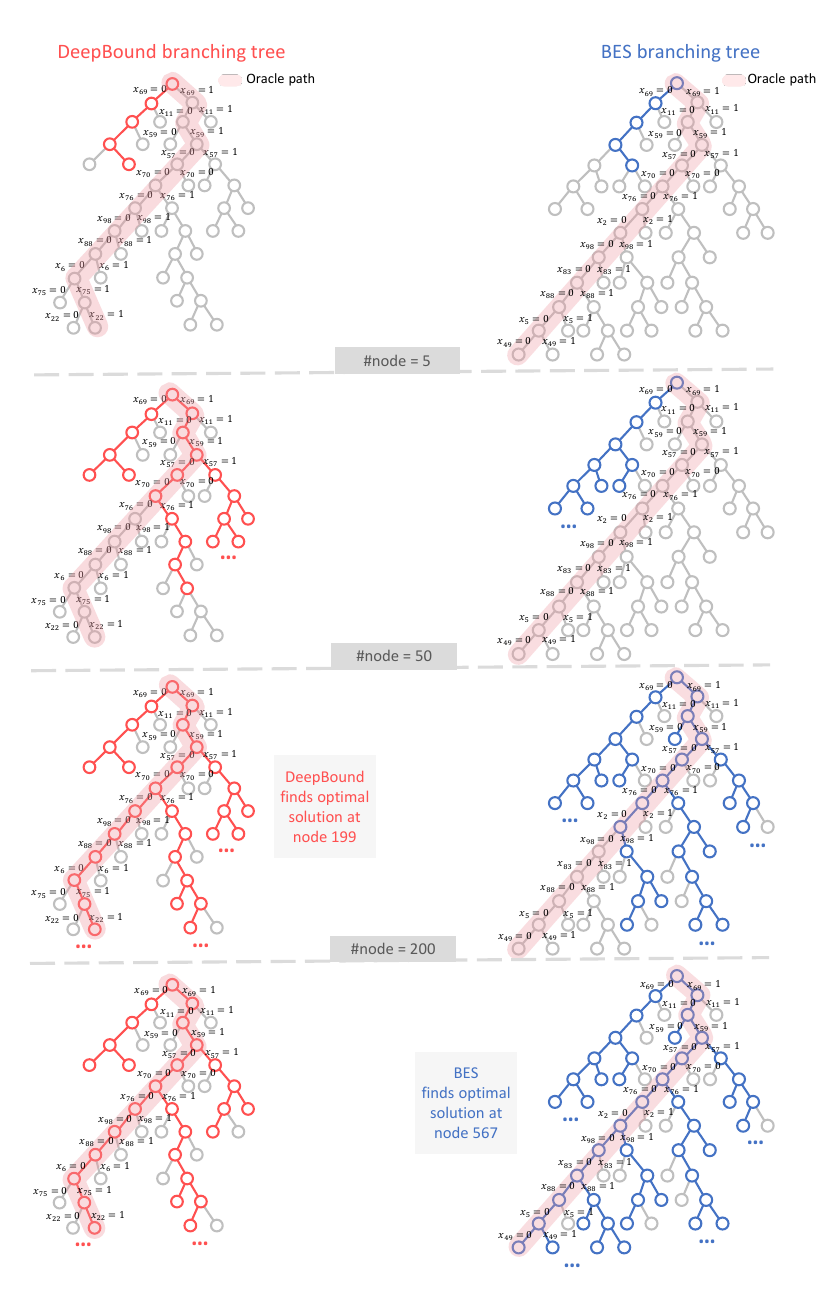}
\caption{\textbf{DeepBound accelerates the solving of MILP problems}. Comparison of the branch-and-bound trees between DeepBound and BES when solving identical 400$\times$100 capacitated facility location problem. 
}
\label{Tree-compare}
\end{figure}

\par

\newpage
\section{Definitions}\label{secA2}

\subsection{Node selection rules}\label{subsec4-2}
In the process of node selection, finding the node with the optimal solution as soon as possible can tighten the primal-bound of the MILP's branch-and-bound search tree rapidly, which reduces the size of the search tree by pruning unpromising nodes \cite{junger200950, achterberg2009scip}. Although limited by primal-bound and dual-bound, extremely large number of nodes may be created in the branch-and-bound solving process of practical problems, which consist of thousands or even millions of constraints and variables. In such conditions, the node selection rules are on a mission to guide the branch-and-bound algorithm to nodes that are likely to contain better feasible solution or better LP relaxation which can help to tighten global dual-bound \cite{achterberg2013mixed}. For the former target, the depth-first-search (DFS) rule greedily selects the child of current solving node, by which may find an integer feasible solution as soon as possible \cite{achterberg2009scip}. For the latter target, the best-first-search (BFS) rule is designed to select node with the worst LP relaxation solution to be the next solving node, which can improve global dual-bound at every node selection step \cite{achterberg2009scip}. However, on the one hand, although DFS can obtain integer feasible solution very early, it completely neglects the tightening of dual-bound, which results in tediously long solving time. On the other hand, although BFS efficiently reduces redundant LP relaxation search space, it cannot promote the tightening of global primal-bound. In order to take care of both two goals, best-estimate-search (BES) rule will calculate the estimate score of every node, which combines information about the dual bound with the information about integrality of the LP solution. It will select the node with the lowest score to solve, which means higher possibility to improve global dual-bound and to find a better integer feasible solution. In practice, in order to speed up the search for an integer feasible solution, BFS rules or BES rules will be combined with the plunging or diving technology, which quickly explores the children of the current node, trying to find a better integer feasible solution \cite{achterberg2009scip}.

\subsection{Branching rules}\label{subsec4-3}
The target of branch variable selection algorithm is to select the branch variable that can significantly improve the LP relaxation of sub-problems, by which can tighten the dual bound of the branch-and-bound tree as soon as possible to reduce the calculation of redundant nodes \cite{achterberg2005branching}. To this end, full-strong branching (FSB) is designed to calculate the LP relaxation of every potential branching variable to test which one gives the best improvement in the dual-bound before actually branching on any of them. Although FSB can provide a gold standard to branching variable selection, its unbearably time consumption makes it scarcely practicable to solve large-scale MILP problems. In order to improve the computing efficiency, pseudocost branching is proposed to utilize the branching history information as an indication of how a branching variable improved the global dual bound before this selection \cite{achterberg2009scip}. By combining branching history and measurement if infeasibility, pseudocost branching dramatically speeds up branching calculation with the cost of large number of redundant nodes because of sparse branching history information at the beginning of branch-and-bound solving process. To overcome this disadvantage, hybrid branching and reliability branching (RPB) \cite{achterberg2009hybrid} combines FSB branching and pseudocost branching by using the former at the early stage, which provides informative branching history to the latter. In practice, RPB is the default branching rule of SCIP  \cite{achterberg2009scip} to trade off between LP relaxation efficiency and the size of branch-and-bound search tree.

\newpage
\section{Methods}\label{secA3}

\subsection{Evaluation data sets}\label{subsec4-4}

Our dataset consists of benchmarks for three different NP-hard MILP problems: set covering, combinatorial auctions, and capacitated facility location. For the set covering problem, we follow the settings described by Balas and Ho \cite{balas1980set}, generating instance sets with 1,000 columns (i.e., variables) and 500 (easy), 1,000 (medium), and 2,000 (hard) rows (i.e., constraints). Unlike the model trained on instances with 500 rows in Gasse et al. \cite{gasse2019exact}, we train and test on a set covering instance with 1,000 rows because SCIP was unable to generate sufficient node data when solving instances with 500 rows. The combinatorial auction instances were generated following the procedure by Leyton-Brown et al. \cite{leyton2000towards}, producing instance sets of varying difficulty: 500 bids with 100 items (easy), 1,000 bids with 200 items (medium), and 1,500 bids with 300 items (hard). Similar to the set covering problem, to generate sufficient node training data, we train our model on instances of 1,000 bids with 200 items and evaluated across all three instance scales. For the capacitated facility location problem \cite{cornuejols1991comparison}, we generated instance sets with 100, 200, and 400 customers. The model was trained on a training set of 200-customer instances and tested on all three difficulty levels. For all three problems, the training set consists of 5,000 instances, while the test sets for each difficulty level contain 100 generated problem instances.

\subsection{Training data generation}\label{subsec4-5}

To collect the training trajectories of oracle nodes, we solved the original MILP problem twice using SCIP with the same FSB branching rule. The first run was used to obtain the optimal solution to the original problem, and the second run labeled the oracle nodes using the known optimal solution. We employed the FSB branching variable selection rule because its branching decisions are independent of node selection history, unlike the RPB branching rule used in previous method. If the RPB rule were used, as in He et al. \cite{he2014learning}, the branching history would heavily influence the heuristic decisions, leading to entirely different branching variable selection results during the second solve, thus introducing inconsistencies. After discovering the optimal solution, SCIP further expands the remaining nodes under global primal and dual bound constraints to prove optimality. All nodes generated during this phase are labeled as non-oracle nodes.

To mitigate the impact of the imbalance between positive and negative samples in the training data, we introduce a pairwise training protocol and a ranking-based learning method. On one hand, although oracle nodes are rare in the branch-and-bound search tree, pairing them with non-oracle nodes randomly selected from the same priority queue forms training pairs. This expands the number of effective node pairs and improves the utilization of positive samples in the branch-and-bound tree. On the other hand, pairing one positive sample node with all negative sample nodes in the same priority queue for training creates a data combination that helps the model learn the differences between positive and negative samples.

\begin{algorithm}
\caption{Training Data Generation}
\begin{algorithmic}[1]
    \Require MILP problem instance $Q$, initial priority queue $\mathcal{P} = \{\mathcal{N}_0\}$
    \Ensure Training dataset $\mathcal{D}^{(Q)}$
    \State Solve $Q$ via branch-and-bound solver
    \If{$S^*$ found during presolve}  \Comment{$S^*$: optimal solution}
        \State $\mathcal{D}^{(Q)} \gets \emptyset$
    \Else
        \State Let $\mathcal{N}^*$ denote the node where $S^*$ is discovered 
        \State Let $PathNodes$ be nodes on the path from $\mathcal{N}_0$ to $\mathcal{N}^*$
        \State Mark $\mathcal{N}^*$ and $PathNodes$ as $\mathbb{B}$ \Comment{$\mathbb{B}$: BPB-Nodes}
        
        \State Re-solve problem $Q$
        \While{$\mathcal{P} \neq \emptyset$}
            \If{$\exists \mathcal{N}_i \in \mathbb{B} \cap \mathcal{P}$}
                \State $\mathcal{D}^{(Q)} \gets \mathcal{D}^{(Q)} \cup \{[(\mathbf{x}_i, \mathbf{y}_i),(\mathbf{x}_j, \mathbf{y}_j)]\}_{\forall \mathcal{N}_j \in \mathcal{P} \setminus \{\mathcal{N}_i\} }$
                \Comment{$\mathbf{x}_i$: features, $\mathbf{y}_i$: labels}
            \EndIf
            \State Select $\mathcal{N}^{\prime} = \argmax_{\mathcal{N} \in \mathcal{P}} score(\mathcal{N})$ \Comment{$score(\cdot)$: scoring function of solver}
            \State Branch $\mathcal{N}^{\prime}$ into two children $\{\mathcal{N}^{\prime}_{1}, \mathcal{N}^{\prime}_{2}\}$
            \State $\mathcal{P} \gets (\mathcal{P} \setminus \{\mathcal{N}^{\prime}\}) \cup \{\mathcal{N}^{\prime}_{1}, \mathcal{N}^{\prime}_{2}\}$
        \EndWhile
    \EndIf
    \State \Return $\mathcal{D}^{(Q)}$
\end{algorithmic}
\end{algorithm}

\subsection{Model architecture}\label{subsec4-6}

\begin{figure}[!thbp]
\centering
\includegraphics[width=1.0\columnwidth]{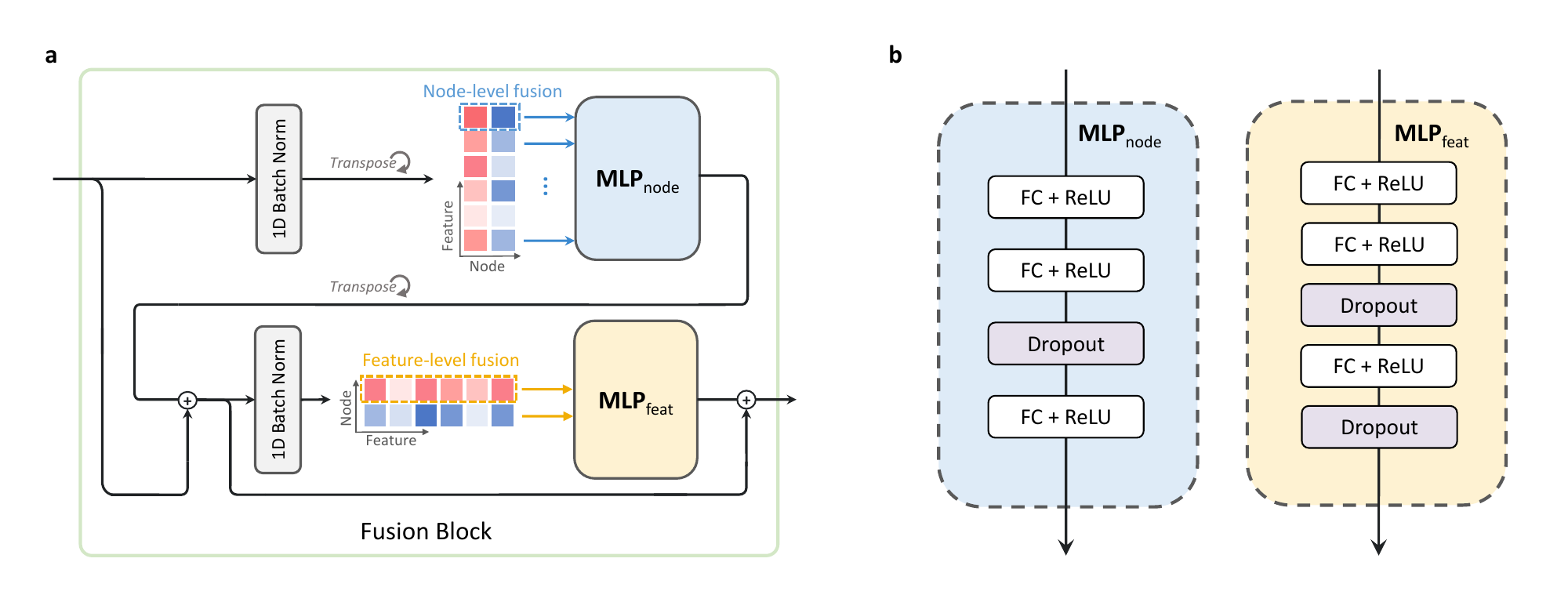}
\caption{\textbf{Fusion block in DeepBound}. \textbf{a,} The fusion block schematic demonstrates multi-level feature integration for node pair, where $MLP_{node}$ performs node-level feature aggregation, while $MLP_{feat}$ enables feature-level fusion. \textbf{b,} Detailed implementations of the core components $MLP_{node}$ and $MLP_{feat}$.}
\label{fig-fusion-block}
\end{figure}

We use a multi-level feature fusion neural network (fusion model) to learn the representation of node features, with each fusion module consisting of several concatenated fusion blocks. We notice that previous work utilized graph neural networks or MLPs designed to encode the features or representations of individual nodes without focusing on information aggregation and comparison across multiple nodes. Therefore, the fusion block includes two modules: a node-level feature fusion module and a feature-level feature fusion module. These modules are constructed using MLPs with different structures, each introducing varying numbers of dropout layers to prevent overfitting and using residual connections to achieve feature fusion.

Additionally, since we use the DeepBound model to encode and score node features, the robustness of the model is critical. We integrate ensemble learning mechanisms into the training of the DeepBound model. The complete DeepBound node selector consists of M independent fusion models, trained by ensemble learning techniques. Inspired by the bagging method, we partition the training data into M subsets, each subset is used to train one independent fusion model. The training hyperparameters are kept consistent in all models. When evaluating the model performance, the final score of the ensemble model is the average of the scores from the M fusion models.

To better assess the training results, we apply K-fold cross-validation on the training set. Different partitions of the training data are used to train the ensemble models. The best-performing ensemble model of this process is selected as the final test model. This approach helps mitigate the issue of high variance from individual fusion model's outputs and improves the stability of the node selector.

\subsection{Input features}\label{subsec4-7}

We use various features of nodes to form a vector that describes the node information. The intrinsic features of a node primarily include the node lower bound, which is the value of the objective function after the node's linear relaxation, and the Node Estimate, which is the estimated value of the node derived from the Best First Search (BFS) rule. Additional features include the node depth and node type (whether it is a child, sibling, or another leaf node of the current solving node). Furthermore, we incorporate features describing the current state of the branch-and-bound tree, such as the global upper and lower bounds, into the input feature vector.

\subsection{Training loss}\label{subsec4-8}


We train the DeepBound model using a mean squared error (MSE) loss between predicted scores and ground-truth labels. The loss function is defined as:
\begin{equation}
\mathcal{L}(\theta) = \frac{1}{n} \sum_{i=1}^{n} \left( p_{\theta}(f_{i}) - y_{i} \right)^2
\end{equation}
where:
\begin{itemize}
    \setlength\itemsep{0.1em}
    \item $n$ denotes the number of nodes in a training batch
    \item $f_i$ represents the feature vector of the $i$-th node
    \item $y_i$ indicates the ground-truth score of the $i$-th node
    \item $p_{\theta}$ denotes the scoring function of the DeepBound model with parameters $\theta$
\end{itemize}
This objective function directly optimizes the model's ability to accurately regress node scores, aligning predictions with the oracle ranking labels.

\subsection{Model inference}\label{subsec4-9}

During inference, we replace the node selection module in the SCIP solver with the average node score of multiple fusion models integrated in the DeepBound model. When solving mixed-integer programming problems via the branch-and-bound algorithm, the solver selects the highest scoring node from the priority queue containing all unsolved nodes to branch on. The feature vectors of the two newly generated nodes from branching are input as pairs into the DeepBound model. These feature vector pairs are encoded by $M$ independent fusion models, generating $M$ sets of scores, which are then averaged to obtain the final output score for the node pair.

This score is subsequently used by the solver to rank the two new nodes when inserting them back into the priority queue, ensuring that all unsolved nodes are ordered in descending score. The branch-and-bound solver, combined with the DeepBound model, continues this process until all unsolved nodes in the priority queue have either been solved or pruned according to the global upper bound, ultimately yielding the exact solution to the original mixed-integer programming problem.

\end{appendices}

\end{document}